\documentclass[]{antgroup}
\PassOptionsToPackage{dvipsnames}{xcolor}
\PassOptionsToPackage{numbers,compress}{natbib}
\usepackage{antgroup}
\usepackage{amsmath,amssymb}
\usepackage[most]{tcolorbox}
\usepackage{fontawesome5}
\usepackage{lipsum}
\tcbuselibrary{breakable}
\usepackage{siunitx}
\usepackage{graphicx}
\usepackage{tikz}
\usepackage{todonotes}
\usepackage{multirow}
\usepackage{cleveref}
\usepackage{multicol}
\usepackage{array}
\usepackage{bm}
\usepackage{enumitem}
\usepackage{algorithm}
\usepackage{algpseudocode}
\usepackage{tabularx}
\usepackage{bbm}
\usepackage{makecell}
\usepackage{subcaption}
\usepackage{colortbl}
\DeclareCaptionLabelFormat{fullfig}{\figurename~\thefigure#2}
\usepackage[normalem]{ulem}
\useunder{\uline}{\ul}{}
\usepackage{fontawesome5}
\newcommand{\corrauthor}{\textsuperscript{\faIcon[regular]{envelope}}}
\newcommand{\corrmark}{\faIcon[regular]{envelope}}

\usepackage{xcolor}
\definecolor{firstcolor}{HTML}{C3423F}
\definecolor{secondcolor}{HTML}{2A4B8C}

\title{UpLoop: Upcycling Large Language Models with Gated Loop}

\author{Tieyuan Chen$^{1,2*}$$^\dag$, Huabin Liu$^{1}$$^\dag$\corrauthor, Yingtao Zhang$^{1}$, Shijie Li$^{2}$, Tengyu Pan$^{1,3}$, Weiyao Lin$^{2}$, Jianguo Li$^{1}$\corrauthor}

\affiliation{$^1$~Inclusion AI, Ant Group\quad}
\affiliation{$^2$~Shanghai Jiao Tong University\quad}
\affiliation{$^3$~Tsinghua University}

\begin{document}

\maketitle

\begin{abstract}
Looped large language models trade parameter storage for recurrent computation, but most are trained from scratch. In practice, pretraining with looped depth is expensive, making post-training upcycling of a pretrained non-looped LLM an attractive alternative.
We introduce \textbf{UpLoop}, a post-training recipe based on \emph{layer-wise} recurrence. Instead of looping a whole model or a middle block, UpLoop loops each selected layer locally, preserving the pretrained depth hierarchy and avoiding feature-space mismatch. To make loop execution useful, we add hidden-wise gates on the attention and MLP residual branches, so later loops act as progressive refinements rather than redundant rewrites.
Empirically, UpLoop breaks the commonly observed saturation at two loops ($R=2$) in prior looped models. Starting from Qwen3-4B Base and applying SFT, UpLoop improves the average accuracy across 20 benchmarks by 6.17\% at $R=3$. We use $R=3$ as the main setting for the best accuracy--cost trade-off, while both native and extrapolated $R=4$ yield further gains.  
UpLoop also generalizes to the MoE backbone Ling-3.0 Tiny, improving the 17-benchmark average by 4.54\%. At inference time, UpLoop further supports practical acceleration via self-speculative decoding and KV-Cache or recurrent-state sharing, achieving over $2\times$ speedup.
\end{abstract}

\section{Introduction}
\label{sec:introduction}
\footnotetext{$^*$Work done at Ant Group. \corrmark\ Corresponding Authors. $^\dag$Core Contributors. Codes and model will be released soon.}
\footnotetext{Email:\{tieyuanchen@sjtu.edu.cn, luobin.lhb@antgroup.com, lijg.zero@antgroup.com\}}
Large language model capability is commonly scaled by increasing parameters, data, or generated reasoning tokens. Looped Transformers provide a different axis: they reuse the same parameters multiple times in hidden space, trading parameter storage for additional computation. Recent work suggests that such looped computation can introduce a favorable bias for iterative reasoning and latent computation \citep{saunshi2025reasoning,ouro,fu2026simply,deng2026lt2,kirk2026stars}. As the area has developed, the focus has broadened from small-scale demonstrations of latent reasoning to questions of scaling, stability, efficiency, and deployment in realistic language model \citep{lee2026sparse,dnd,gao2026loop,SMELT,jeddi2026loopformer,MELT,fein2026solve,fang2026loopq}.

In practice, however, a more relevant question is whether a strong pretrained \emph{non-looped} LLM can be \emph{upcycled} into a useful recurrent-depth model only through continued pretraining or supervised fine-tuning, rather than trained as a looped model from scratch.
This question matters because pretraining with looped depth is often substantially more expensive. Looping during pretraining directly multiplies recurrent computation, increases optimization complexity, and complicates stability and systems efficiency \citep{yang2026loopcoder,zeitoun2026hyperloop,ouro,SMELT}. 
In many realistic settings, it is therefore preferable to start from a large-scale pretrained LLM and ask whether the same post-training recipe can further improve the model through hidden-state recurrence. Our goal is precisely this form of \emph{post-training upcycling}.

However, post-training upcycling is not straightforward. After large-scale pretraining, an LLM develops a stable depth-wise feature hierarchy: shallow layers process embedding-side features, middle layers transform and compose them, and deeper layers produce representations closer to readout space~\citep{koishekenov2025encode,SMELT}. 
This makes recurrence schemes common in native looped LLMs trained from scratch---especially \emph{model-wise} and \emph{block-wise} looping---poorly suited to upcycling, because they feed late-stage representations back into shallower interfaces that were not pretrained to process them.
We therefore adopt the \emph{layer-wise loop} of \emph{Loop the Loopies} \citep{gao2026loop}: in \textbf{UpLoop}, each selected layer performs looped visits locally before passing the hidden state onward, thereby preserving the pretrained depth hierarchy.

Moreover, preserving the pretrained depth hierarchy alone is not sufficient to make recurrent computation useful. Prior looped models exhibit a recurring empirical limitation: the most reliable gains are often obtained at two loops, while increasing the loop count beyond $R=2$ frequently brings little benefit or even degrades performance \citep{SMELT,yang2026loopcoder,gao2026loop, LoopMDM,schwethelm2026much,prairie2026parcae,labovich2026stability,kohli2026loop,oncescu2026recurrent}. 
Here and throughout the paper, $R$ denotes the number of local loop executions performed by each selected looped layer. 
We refer to a model instantiated with loop count $R=i$ as an ``$R=i$ model,'' meaning that each selected looped layer is executed for $i$ local rounds indexed by $r\in\{1,\ldots,i\}$.
The case $r=1$ corresponds to the original pretrained forward pass of that layer, while the rounds $r=2,\ldots,i$ are additional recurrent refinement passes that reuse the same parameters.
A central goal of UpLoop is to overcome the ``$R=2$ ceiling'' in post-training loop upcycling.

To make repeated execution useful, we therefore modulate the extra loops with learned hidden-wise gates on the attention and MLP residual writes. This design is inspired by stage-dependent modulation in Diffusion Transformers~\citep{dit}: different refinement steps need not write equally. 
In UpLoop, the modulation is implemented as learned gate vectors that are separately parameterized for each layer, loop, and residual branch. Thus, although parameters are shared across loops, later loops can still apply different residual write strengths at different depths and hidden channels. 
Concretely, we keep the first loop identical to the pretrained forward pass and apply these layer- and loop-specific gates only to later loops. 
This preserves the original pretrained computation, while allowing subsequent loops to inject different amounts of new information. 
The gate therefore serves as a write controller that reduces redundancy, suppresses unnecessary overwriting, and enables controlled progressive refinement. 

We further apply intermediate supervision and an improvement objective so that each recurrent depth remains decodable and later loops refine rather than degrade earlier ones. This is motivated by prior findings that looped models can otherwise suffer from drift and poorly controlled latent trajectories \citep{chen2026loop,RLTT,blayney2026mechanistic,labovich2026stability}.

Empirically, starting from Qwen3-4B Base~\citep{qwen} and conducting supervised fine-tuning (SFT) on about 13B tokens, UpLoop breaks the commonly observed $R=2$ saturation pattern: the average accuracy across 20 benchmarks improves from 51.90 for the non-looped baseline to 55.56 at $R=2$ and further to 58.07 at $R=3$. We use $R=3$ as our main setting because it offers the best accuracy--cost trade-off. Notably, $R=4$ remains beneficial: a natively trained $R=4$ model reaches 58.70 on average, and inference-time extrapolation from trained $R=3$ to test-time $R=4$ also improves the average to 58.55.
UpLoop also generalizes to the MoE backbone Ling-3.0 Tiny, improving the 17-benchmark average from 50.97 to 55.51 at $R=3$. At inference time, UpLoop further supports practical acceleration via self-speculative decoding together with backbone-specific cross-loop sharing, achieving up to $2.05\times$ speedup on Qwen3-4B and $2.30\times$ on Ling-3.0 Tiny.

Overall, our contributions are threefold:
\begin{itemize}[leftmargin=1.5em]
    \item We introduce \textbf{post-training loop-depth upcycling}, showing that a \emph{non-looped} pretrained LLM can be converted into a useful recurrent-depth model through only SFT.
    \item We propose \textbf{UpLoop}, a layer-wise recurrent architecture with gated attention/MLP residual writes and depth-aware supervision, which makes loop depths beyond $R=2$ effective.
    \item UpLoop improves the 20-benchmark average by \textbf{6.17\%} on Qwen3-4B and the 17-benchmark average by \textbf{4.54\%} on Ling-3.0 Tiny; with self-speculative decoding and cross-loop sharing, UpLoop-Qwen3-4B and UpLoop-Ling-3.0 Tiny achieve up to $\bf{2.05\times}$ and $\bf{2.30\times}$ inference speedup, respectively.
\end{itemize}

\section{Related Work}
\label{sec:related}

\paragraph{Training regime and recurrence granularity in looped language models.}
Recent looped and recurrent-depth language models repeatedly apply a shared layer or block to increase effective depth under parameter sharing, motivated by the idea that hidden-space recurrence can support iterative latent computation and reasoning \citep{saunshi2025reasoning,ouro}. Most prior work focuses on \emph{from-scratch} looped pretraining, where recurrence is typically organized over a shared whole model or middle stack, i.e., \emph{model-wise} or \emph{block-wise} looping \citep{saunshi2025reasoning,ouro,prairie2026parcae,gao2026loop,SMELT}. In contrast, \emph{Loop the Loopies} \citep{gao2026loop} proposes \emph{layer-wise} looping, which keeps recurrent updates within a more limited feature-space neighborhood and is better suited to upcycling pretrained models.

In parallel, several works have explored post-hoc conversion of pretrained non-looped LLMs. \emph{LoopUS} \citep{LoopUS} turns a pretrained model into an encoder--reasoning--decoder architecture through continued training, using a shared middle reasoning block together with selective gating. However, LoopUS operates at the \emph{block} level; its design can be less general across backbones: the choice of which intermediate block should serve as the loop module typically depends on model-specific tuning, and its gating acts at a relatively coarse granularity compared with finer layer-wise modulation. Similarly, \citep{Retrofitted} retrofit pretrained models into a prelude--recurrent-block--coda architecture by pruning intermediate layers and inserting linear adapters; however, it necessitates a costly CPT phase (e.g., 50B tokens) to recover base capabilities.
Adopting \emph{model-wise} looping, \emph{Think-at-Hard (TaH)} \citep{tah} and \emph{TaH2} \citep{tah2} post-train pretrained LLMs to loop over the entire layer stack, pairing the backbone with a learned state-updater MLP and a token-level iteration decider. However, because model-wise looping feeds late-stage representations back into the shallowest layer, it inherently suffers from severe feature-space mismatch, necessitating auxiliary input-injection modules, duo-causal attention masks, and dynamic token halting.
\emph{Training-Free Looped Transformers} \citep{chen2026training}, in contrast, introduces a purely inference-time loop wrapper over a frozen checkpoint, repeatedly applying a contiguous mid-stack block without any additional training. Our setting differs from both: we neither restructure the model into a new encoder--reasoning--decoder form nor treat recurrence as a training-free inference patch. 
Instead, we upcycle pretrained LLMs with \emph{layer-wise} local recurrence, which better preserves pretrained depth hierarchy and limits feature-space mismatch.

\paragraph{Architectural and optimization mechanisms for effective loop.}
A second major theme in the looped-model literature is how to make additional recurrent steps useful rather than redundant or unstable. Prior work has explored architectural modifications and training objectives to improve recurrent information flow, stability, and controllability \citep{LoopUS,chen2026loop,RLTT,blayney2026mechanistic,labovich2026stability,zeitoun2026hyperloop}. 
However, current studies suggest that loop depth does not scale easily in practice: useful gains often peak around two recurrent steps, and increasing the number of training-time loop steps frequently yields little or no further improvement \citep{SMELT,yang2026loopcoder}. 
We argue that this empirical ceiling is not necessarily a fundamental limit of recurrence itself, but may instead reflect insufficient control over recurrent writes. 
While concurrent work \citep{li2026scaling} shares the motivation of preventing destructive overwrites, it relies on block-level scalar attention pooling with temporal decay to scale from-scratch 7M puzzle reasoners over thousands of steps. 
In contrast, UpLoop is tailored for upcycling 4B-Level pretrained LLMs: rather than coarse scalar state averaging across blocks, we inject channel-wise gates directly into each layer's internal attention and MLP branches, breaking the $R=2$ ceiling across 20 general LLM benchmarks. Latent-reasoning work~\citep{geiping2026scaling} also scales loop depth by repeatedly conditioning on the initial input, effectively extending implicit reasoning before next-token prediction; despite the different setting, it shares our core intuition of balancing preserved initial information against later recurrent updates.

\section{Method}
\label{sec:method}

UpLoop is built around three components. First, we adopt a \emph{layer-wise loop}. 
Second, we add \emph{gated residual writes} on the extra loops, using a gate to modulate how much new information is injected. 
Third, we introduce a \emph{depth-aware objective} that keeps intermediate loops decodable and encourages later loops to make progress.

\subsection{Layer-Wise Loop}
\label{sec:method_local_loop}

A common looped design is to loop the whole model or a middle stack of layers. This works naturally when the looped model is trained from scratch, but is less well matched to post-training. 
In a pretrained decoder, earlier and later layers are not interchangeable: shallow layers are optimized to consume embedding-side features, while deeper layers operate on more task-ready representations~\citep{koishekenov2025encode}. Refeeding the deep-layer representations back into shallow layers therefore introduces a feature-space mismatch.

Similar to \citep{gao2026loop}, UpLoop avoids this mismatch by making looping \emph{local} to each layer; each selected layer executes several local loops to refine its own representation before passing the hidden state forward.

Let $L_1,\ldots,L_N$ denote the pretrained Transformer layers, and let $h_0$ denote the hidden state entering the first selected looped layer.
For layer $i\in\{1,\ldots,N\}$ and local loop index $r\in\{1,\ldots,R\}$, we define
\begin{equation}
h_{1,0} = h_0,
\qquad
h_{i,0} = h_{i-1,R}\ \text{for } i=2,\ldots,N,
\qquad
h_{i,r} = L_i(h_{i,r-1}), \ \ r=1,\ldots,R.
\label{eq:layerwise_loop_basic}
\end{equation}
Thus, $h_{i,0}$ denotes the hidden state entering physical layer $i$, and $h_{i,R}$ is the output of that layer after $R$ local loops, which is then passed to layer $i+1$. The parameters of $L_i$ are shared across the $R$ local loops of the same physical layer $i$.
Under this convention, the case $r=1$ corresponds to the original pretrained single-pass computation of layer $i$, while the rounds $r=2,\ldots,R$ are additional recurrent refinement steps that reuse the same layer parameters. This local recurrence preserves the pretrained depth hierarchy, because each repeated update remains within the feature space of the same physical layer.

\subsection{Gated Residual Writes for Scaling Looped Depth}
\label{sec:method_gate}

Simply repeating a layer several times does not guarantee useful extra computation.
For post-training upcycling, two issues are central: full-strength extra loops can destabilize the pretrained representation, and without a small-step regime, repeated loops may fail to produce progressive refinement.
We therefore want additional loops to act as \emph{controlled refinement}: stable at each loop, yet cumulatively depth-like across loops.

To this end, UpLoop modulates later loops with gated residual writes. For each looped layer $i$ and each later local loop $r\in\{2,\ldots,R\}$, we introduce branch-specific hidden-wise gate vectors
\begin{equation}
g^{b}_{i,r}=\sigma\!\bigl(s\,\theta^{b}_{i,r}\bigr)\in(0,1)^d,
\qquad
b\in\{\mathrm{attn},\mathrm{mlp}\},
\label{eq:gate_def}
\end{equation}
where $\theta^{b}_{i,r}\in\mathbb{R}^d$ is a learned raw gate vector, $\sigma(\cdot)$ is the sigmoid function, $d$ is the hidden dimension, and $s$ is a fixed logit scale. Thus, the raw gate parameters are unconstrained, while the effective residual write coefficients always lie in $(0,1)$.
For the first local loop $r=1$, we keep the original pretrained layer computation unchanged. For each later loop $r>1$, we place hidden-wise gates on the attention and MLP residual branches:
\begin{align}
\tilde h_{i,r}
&=
h_{i,r-1}
+
g^{\mathrm{attn}}_{i,r}\odot
\operatorname{Attn}_i\!\bigl(\operatorname{Norm}_i(h_{i,r-1})\bigr),
\\
h_{i,r}
&=
\tilde h_{i,r}
+
g^{\mathrm{mlp}}_{i,r}\odot
\operatorname{MLP}_i\!\bigl(\operatorname{Norm}'_i(\tilde h_{i,r})\bigr).
\label{eq:gated_layer_update}
\end{align}
The gates are hidden-channel-wise vectors, allowing different hidden dimensions to take different residual write strengths at different layers, loops, and residual branches. In this way, repeated execution is parameterized as loop-specific, channel-wise residual refinement rather than uniform full-strength rewriting.

To make the role of the gate explicit, it is convenient to introduce an abstract residual form. Let $F_i(\cdot)$ denote the effective residual transformation of layer $i$ after composing the normalized attention and MLP residual writes into a single update map. To unify the notation across all local loops, we define
\begin{equation}
D_{i,1} := I_d,
\qquad
D_{i,r} := \operatorname{Diag}(g_{i,r}) \in [0,1]^{d\times d},
\ \ r=2,\ldots,R,
\label{eq:gate_diag_def}
\end{equation}
where $I_d$ is the $d\times d$ identity matrix, and $g_{i,r}\in[0,1]^d$ denotes an abstract hidden-wise gate vector for loop $r$ after suppressing the attention/MLP branch split for notational simplicity. Then each local loop can be written as
\begin{equation}
h_{i,r}= \Phi_{i,r}(h_{i,r-1})
:= h_{i,r-1}+ D_{i,r}\,F_i(h_{i,r-1}),
\qquad r=1,\ldots,R.
\label{eq:gated-recur-abstract}
\end{equation}
Unrolling the recurrence gives
\begin{equation}
h_{i,R}
=
h_{i,0}
+
\sum_{r=1}^{R}
D_{i,r}\,F_i(h_{i,r-1}).
\label{eq:gated-recur-unroll}
\end{equation}
Thus, the effective computation is not a single repeated overwrite, but an accumulation of loop-specific residual refinements.
This perspective is useful because it recasts looping as a sequence of gated residual steps, rather than repeated full-strength rewrites of the hidden state.
This abstraction also clarifies why gating improves stability.
Assume locally that $F_i$ is $L_F$-Lipschitz, and that the gate matrix satisfies
\begin{equation}
\|D_{i,r}\|_{\mathrm{op}} \le \alpha_{i,r}\le 1.
\label{eq:gate-local-assump}
\end{equation}
Then for any two nearby states $x,y$, one loop satisfies
\begin{align}
\|\Phi_{i,r}(x)-\Phi_{i,r}(y)\|
&\le
\|x-y\|
+
\|D_{i,r}\|_{\mathrm{op}}\,\|F_i(x)-F_i(y)\| \nonumber\\
&\le
\bigl(1+\alpha_{i,r}L_F\bigr)\|x-y\|.
\label{eq:gated-one-step-bound}
\end{align}
Compared with the ungated case, whose corresponding factor is $1+L_F$, the gate explicitly shrinks the residual contribution through $\alpha_{i,r}$.
Thus, when $\alpha_{i,r}<1$, perturbations are amplified less aggressively across loops.
Intuitively, the gate moves looping toward a better-conditioned small-step regime.

This small-step regime is also what connects gated looping to \emph{effective depth}. A standard non-looped model with $N$ stored layers performs $N$ feature transformations. A layer-wise looped model with $R$ loops still stores only $N$ layers, but it executes approximately $RN$ transformation steps. Of course, this does not make it equivalent to an untied $RN$-layer network, since parameters are shared across loops. However, when each residual write is small, the multi-loop layer admits a first-order expansion that behaves like a depth-expanded residual stack with shared weights and loop-dependent coefficients.

Concretely, suppose each gated residual step is uniformly small:
\begin{equation}
\|D_{i,r}F_i(h_{i,r-1})\| \le \epsilon,
\qquad r=1,\dots,R,
\label{eq:small-step-assump}
\end{equation}
and suppose $F_i$ is locally smooth around $h_{i,0}$. Since each loop changes the state by at most $\epsilon$, we have
\begin{equation}
\|h_{i,r}-h_{i,0}\|=\mathcal{O}(r\epsilon).
\end{equation}
By local smoothness, this implies
\begin{equation}
F_i(h_{i,r-1}) = F_i(h_{i,0}) + \Delta_{i,r},
\qquad
\|\Delta_{i,r}\|=\mathcal{O}(r\epsilon).
\label{eq:first-order-local-expand}
\end{equation}
Substituting \eqref{eq:first-order-local-expand} into \eqref{eq:gated-recur-unroll} gives
\begin{align}
h_{i,R}
&=
h_{i,0}
+
\sum_{r=1}^{R}
D_{i,r}\,F_i(h_{i,0})
+
\sum_{r=1}^{R} D_{i,r}\,\Delta_{i,r}.
\end{align}
Under the small-step assumption, the remainder term satisfies
\begin{equation}
\sum_{r=1}^{R} D_{i,r}\,\Delta_{i,r}
=
\mathcal{O}(R^2\epsilon^2),
\end{equation}
and therefore
\begin{equation}
h_{i,R}
=
h_{i,0}
+
\sum_{r=1}^{R}
D_{i,r}\,F_i(h_{i,0})
+
\mathcal{O}(R^2\epsilon^2).
\label{eq:gated-first-order}
\end{equation}
Thus, to leading order in the small-step regime, the $R$-loop layer behaves like a depth-$R$ residual stack with a shared transformation $F_i$ and loop-dependent diagonal coefficients $D_{i,r}$. In the ungated case, where $D_{i,r}\equiv I$, the same expansion reduces to
\begin{equation}
h_{i,R}
=
h_{i,0}
+
R\,F_i(h_{i,0})
+
\mathcal{O}(R^2\epsilon^2),
\label{eq:ungated-first-order}
\end{equation}
which corresponds to adding the same residual update at full strength across loops. The role of gating is therefore not to enable the expansion itself, but to keep looping in a plausible small-step regime: shrinking recurrent writes limits cumulative drift and turns repeated execution into controlled residual refinement. 
Together, \eqref{eq:gated-one-step-bound} and \eqref{eq:gated-first-order} show why hidden-wise gating is suitable for post-training upcycling: layer-wise looping preserves the pretrained feature hierarchy, while gating reallocates computation to recover part of the benefit of greater depth. 
Full derivations and complementary views are provided in Appendix~\ref{app:theory_gate}.

\subsection{Depth-Aware Supervision}
\label{sec:method_objective}

Even with layer-wise looping and gated residual writes, later loops may fail to improve over earlier ones, and earlier loops may become poorly decodable, effectively serving as temporary scratch space~\citep{ouro}.

Similar to \citep{ouro}, we therefore supervise all loop depths, not just the final one. Let $\ell_r$ denote the token-averaged cross-entropy loss computed from the model output obtained after executing each selected looped layer up to local round $r$, and let $R$ denote the maximum number of loops. We use two auxiliary terms in addition to the final-depth loss.
First, we apply \emph{intermediate supervision}:
\begin{equation}
 \mathcal L_{\mathrm{int}} = \frac{1}{R-1}\sum_{r=1}^{R-1}\ell_r.
\end{equation}
This encourages earlier loops to remain directly readable by the output head, rather than drifting into hidden states that only become useful after several more loops.
Second, we add an \emph{improvement loss}:
\begin{equation}
 \mathcal L_{\mathrm{imp}} = \frac{1}{R-1}\sum_{r=1}^{R-1}
 \max\bigl(0,\ell_{r+1}-\ell_r+m\bigr),
\end{equation}
where $m$ is a margin. This term encourages each successive loop to outperform the previous one by at least a small amount, discouraging stagnant or regressive trajectories.
The full training objective is
\begin{equation}
 \mathcal L = \ell_R + \lambda_{\mathrm{int}}\mathcal L_{\mathrm{int}}
 + \lambda_{\mathrm{imp}}\mathcal L_{\mathrm{imp}}.
\end{equation}

The role of this objective is not merely to improve optimization. It also shapes how looping is used: earlier loops must stay decodable, and later loops are explicitly trained to yield further improvements.

\section{Experiments}
\label{sec:experiments}

\subsection{Setup}
Our primary model uses a maximum local loop count of $R=3$, which we choose as the main setting for the best accuracy--cost trade-off. We also train $R=2$ and native $R=4$ variants, and study inference-time extrapolation from $R=3$ to $R=4$.
We preserve the first and last three decoder layers non-looped.

The gates are parameterized as unconstrained raw vectors and mapped to effective residual coefficients through a sigmoid with logit scale $s=8.0$. We initialize the raw parameters by inverse-sigmoid so that the effective gate value is $0.3$ at initialization. We update all model parameters during SFT.
The training objective combines final-depth CE, intermediate supervision, and improvement loss. In the default run, the intermediate and improvement loss weights are $0.3$ and $5.0$, respectively, with improvement margin $m=0.05$. Training uses BF16, a global batch size of 128, the Muon optimizer~\citep{muon}, and about 3,000 optimization steps.
We further adopt a staged loop-depth curriculum: training starts from the single-pass setting, increases to $R=2$ at step 300, to $R=3$ at step 700, and to $R=4$ at step 1,000.

\subsection{Benchmarks}
We evaluate on 20 benchmarks spanning mathematics, coding, tool use, commonsense reasoning, instruction following, reasoning, and knowledge:
\begin{itemize}[leftmargin=1.5em]
    \item \textbf{Math:} Math500~\citep{math}, GSM-Plus~\citep{li2024gsmplus}, OlympiadBench~\citep{Olympiadbench}, AIME 2025~\citep{aime25}, AIME 2026~\citep{aime26}
    \item \textbf{Code:} HumanEval+~\citep{humaneval}, MBPP+~\citep{mbpp}, LiveCodeBench v6~\citep{livecodebench}, BigCodeBench~\citep{bigcodebench}, CruxEval~\citep{gu2024cruxeval}, Aider~\citep{aider}
    \item \textbf{Reasoning / General:} BFCL-v3~\citep{bfcl}, PIQA~\citep{bisk2020piqa}, MuSR~\citep{musr}, BBH~\citep{bbh}, KorBench~\citep{korbench}, IFBench~\citep{ifbench}, GPQA-Diamond~\citep{gpqa}, MMLU-Pro~\citep{mmlipro}, ZebraLogic~\citep{Zebralogic}
\end{itemize}
All benchmarks are evaluated under sampling-based decoding with at least 4 random samples per example, and we report mean pass@1. For AIME 2025 and AIME 2026, we use 64 samples per example. 

\subsection{Main Results}
\begin{table*}[t]
\centering
\small
\resizebox{0.96\linewidth}{!}{
\begin{tabular}{llcccccc}
\toprule
& & \multicolumn{4}{c}{Comparison under $R=3$} & \multicolumn{2}{c}{Loop count under UpLoop} \\
\cmidrule(lr){3-6} \cmidrule(lr){7-8}
Category & Dataset & Non-looped & Loop w/o gate & UpLoop & $\Delta$ (UpLoop - Non-loop) & $R=2$ & $R=4$ \\
\midrule
\multirow{5}{*}{Math}
& Math500           & 81.40 & 82.20 & \textbf{85.60} & +4.20 & 84.00 & 84.00 \\
& GSM-Plus          & 81.28 & 83.61 & 84.97 & +3.69 & 83.63 & \textbf{85.03} \\
& OlympiadBench     & 53.19 & 56.46 & \textbf{60.81} & +7.62 & 58.22 & 59.00 \\
& AIME 2025         & 26.67 & 31.72 & 36.61 & +9.94 & 33.18 & \textbf{39.53} \\
& AIME 2026         & 26.67 & 26.98 & 36.09 & +9.42 & 32.45 & \textbf{39.27} \\
\midrule
\multirow{6}{*}{Code}
& HumanEval+        & 64.85 & 62.20 & 68.29 & +3.44 & \textbf{70.12} & 73.78 \\
& MBPP+             & 49.74 & 58.85 & 60.32 & +10.58 & 63.76 & \textbf{64.29} \\
& LiveCodeBench v6  & 23.45 & 24.17 & 25.77 & +2.32 & 25.28 & \textbf{26.43} \\
& BigCodeBench      & 29.72 & 30.88 & \textbf{33.28} & +3.56 & 29.47 & 30.88 \\
& CruxEval          & 63.44 & 65.12 & 65.75 & +2.31 & 66.32 & \textbf{66.81} \\
& Aider             & 33.31 & 41.37 & \textbf{43.61} & +10.30 & 37.85 & 37.59 \\
\midrule
\multirow{4}{*}{Reasoning}
& MuSR              & 76.37 & 79.00 & 80.18 & +3.81 & 76.91 & \textbf{83.36} \\
& BBH               & 75.50 & 78.07 & 79.66 & +4.16 & 77.72 & \textbf{79.71} \\
& GPQA-Diamond      & 39.02 & 43.32 & 44.48 & +5.46 & 40.88 & \textbf{46.88} \\
& ZebraLogic        & 55.30 & 51.95 & \textbf{70.03} & +14.73 & 58.42 & 69.33 \\
\midrule
\multirow{5}{*}{General}
& BFCL-v3           & 58.80 & 62.24 & \textbf{66.14} & +7.34 & 64.28 & 63.21 \\
& PIQA              & 80.32 & 83.73 & 84.33 & +4.01 & 83.95 & \textbf{84.60} \\
& KorBench          & 45.60 & 53.04 & 56.56 & +10.96 & 48.72 & \textbf{57.52} \\
& IFBench           & 19.00 & 21.27 & 21.47 & +2.47 & 19.53 & \textbf{23.13} \\
& MMLU-Pro          & 54.37 & 55.81 & 57.54 & +3.17 & 56.44 & \textbf{59.56} \\
\midrule
\multicolumn{2}{l}{Average} & 51.90 & 54.60 & 58.07 & +6.17 & 55.56 & \textbf{58.70} \\
\bottomrule
\end{tabular}%
}
\caption{Main results (\%). The left group compares different strategies under the main $R=3$ setting, while the right group compares different loop counts for UpLoop. Layer-wise loop without gates does not yield a clear gain. This motivates gated UpLoop, which delivers markedly greater improvements. Although performance continues to improve at $R=4$, the gain is relatively small, making $R=3$ the best accuracy--cost trade-off.}
\label{tab:main}
\end{table*}

\Cref{tab:main} shows that layer-wise looping is a reasonable baseline for post-training upcycling. Even without gates, it improves the 20-benchmark average from 51.90 to 54.60, but the gains are not clearly consistent across tasks. 
We therefore further introduce gated UpLoop, which raises the average to 55.56 at $R=2$ and 58.07 at $R=3$, showing that useful loop depth need not saturate at $R=2$ in post-training upcycling. 
A native $R=4$ model reaches 58.70. These results indicate that the gate is crucial for modulating simple layer-wise loop.

Comparing $R=2$ and $R=3$ shows that the gains from looping do not saturate at two loops. We use $R=3$ as the main operating point because it provides the best trade-off between accuracy and cost, although $R=4$ still brings a small further gain. Unless otherwise specified, the following analyses focus on the $R=3$ model.

The improvements are broad rather than isolated. Relative to the vanilla model, the $R=3$ variant improves all 20 measured tasks, with especially large gains on ZebraLogic (+14.73), KorBench (+10.96), MBPP+ (+10.58), Aider (+10.30), AIME 2025 (+9.94), and AIME 2026 (+9.42). These results indicate that the benefit of layer-wise recurrent-depth upcycling is particularly strong on mathematical, coding, and reasoning tasks, while still extending to tool use, commonsense reasoning, instruction following, and knowledge-intensive benchmarks.

\subsection{Progressive Refinement Across Loops}
\label{sec:progressive_refinement}

We ask whether extra local loops do meaningful work.
Specifically, we examine confidence in token prediction and token-level prediction churn. 
As shown in \Cref{tab:confidence_visits}, mean entropy drops from 4.0186 at $r=1$ to 3.3312 at $r=2$ and 1.8674 at $r=3$. Meanwhile, the mean top-1 probability rises.
Thus, later loops do not simply preserve the same predictive distribution; they make the model more decisive at readout.

\begin{table}[t]
\centering
\begin{subtable}[t]{0.58\linewidth}
\centering
\small
\begin{tabular}{lcccc}
\toprule
Metric & $r=1$ & $r=2$ & $r=3$ & $r=1 \rightarrow r=3$ \\
\midrule
Entropy (nats) & 4.0186 & 3.3312 & \textbf{1.8674} & -2.1512 \\
Top-1 probability & 0.1827 & 0.2338 & \textbf{0.4420} & +0.2593 \\
\bottomrule
\end{tabular}
\caption{Confidence statistics across loops.}
\label{tab:confidence_visits}
\end{subtable}
\hfill
\begin{subtable}[t]{0.4\linewidth}
\centering
\small
\begin{tabular}{lc}
\toprule
Transition & Churn rate \\
\midrule
$r=1 \rightarrow r=2$ & 3.11\% \\
$r=2 \rightarrow r=3$ & 3.45\% \\
\bottomrule
\end{tabular}
\caption{All-token top-1 prediction churn across loops.}
\label{tab:decoding_prediction_churn}
\end{subtable}
\caption{Evidence for progressive refinement across loops. Left: later loops become more confident, with lower entropy and higher top-1 probability. Right: later loops still revise a non-trivial fraction of decisions.}
\label{tab:progressive_refinement}
\end{table}

Confidence growth alone, however, could still reflect trivial sharpening of an already fixed choice.
We therefore measure \emph{all-token prediction churn} during greedy decoding.
At each generation step, $r=1$, $r=2$, and $r=3$ receive the same prefix, and we compare their top-1 predictions.
To isolate loop effects from divergence between separately generated sequences, we append the $r=3$ greedy token to a shared history.
As shown in \Cref{tab:decoding_prediction_churn}, later loops remain active at the token level.
So $r=3$ is not merely sharpening an earlier decision; it still revises local decoding choices.

At the same time, churn remains low in absolute terms.
Most positions are preserved across loops, consistent with increasing recurrent stability, while a non-trivial minority are still refined.
Overall, these results support \emph{progressive refinement}: later loops are more confident, but not inert.

\subsection{Gating Converts Repeated Execution into Controlled Refinement}
\label{sec:controlled_updates}

Having shown that later loops remain useful at readout, we next ask why repeated layer reuse does not collapse into redundancy or destructive overwrite.
We test this at three levels: (1) round-wise update statistics, (2) late-layer hidden-state dynamics, and (3) residual-write measurements.

\paragraph{Round-wise statistics show that gating suppresses overwrite without killing refinement.}

\begin{table}[t]
\centering
\small
\resizebox{0.5\linewidth}{!}{
\begin{tabular}{lcc}
\toprule
Metric & With gate & Without gate \\
\midrule
Hidden $\Delta$-RMS $r=1 \rightarrow r=2$ & \textbf{0.47} & 1.19 \\
Hidden $\Delta$-RMS $r=2 \rightarrow r=3$ & \textbf{0.70} & 0.92 \\
$r=2 \rightarrow r=3$ update cosine & \textbf{0.43} & 0.34 \\
$r=2 \rightarrow r=3$ attention KL & \textbf{0.005} & 0.001 \\
$r=2 \rightarrow r=3$ output KL & \textbf{0.31} & 0.23 \\
\bottomrule
\end{tabular}}
\caption{Round-wise update statistics with and without gates. Gating turns repeated loop execution into smaller but more coherent refinement updates, reducing early hidden-state overwrite while preserving meaningful late-round changes in attention and output distributions.}
\label{tab:round_stats}
\end{table}

\Cref{tab:round_stats} shows two key effects.
First, gating sharply reduces the earliest overwrite: the hidden-state $\Delta$-RMS from $r=1$ to $r=2$ drops from 1.19 to 0.47.
This supports the gate's role as a write controller, making the second loop a refinement rather than a near-complete rewrite.
Second, later loops remain non-trivial.
Although gating also reduces the $r=2 \rightarrow r=3$ update magnitude, the update does not vanish.
Instead, the gated model shows more directionally coherent refinement, with update cosine rising from 0.34 to 0.43.
Attention KL and output KL are also larger with gates, indicating that $r=3$ still changes both retrieval patterns and token predictions.

\paragraph{Gating matters most near the top of the decoder.}

\begin{figure*}[t]
    \centering
    \includegraphics[width=0.95\textwidth]{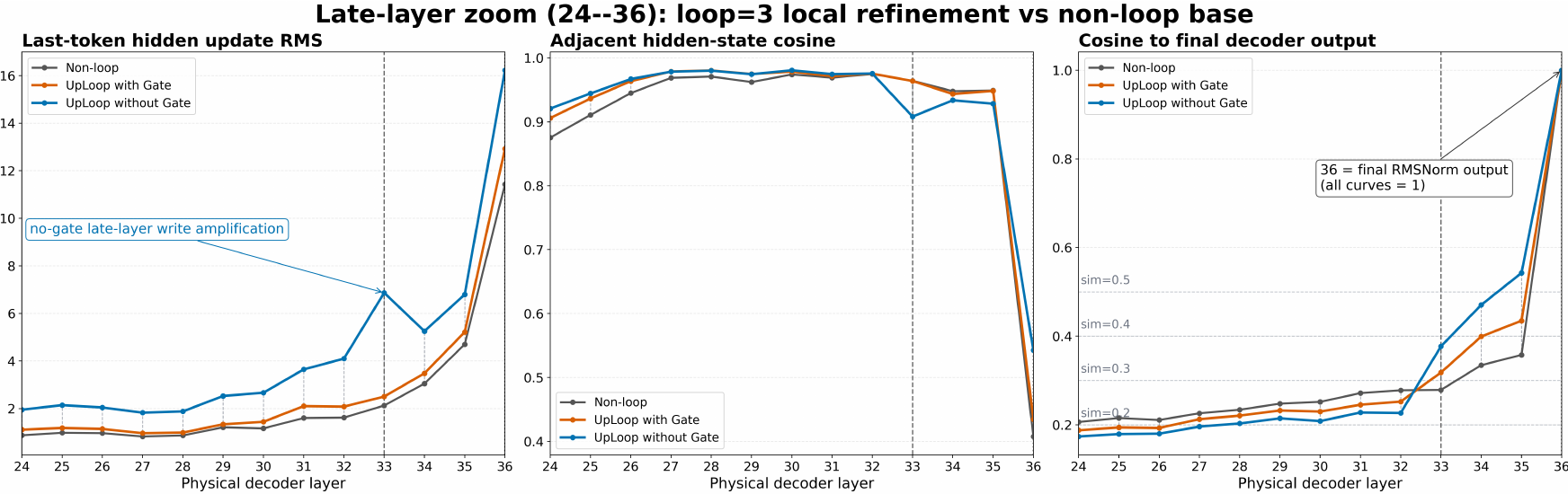}
    \vspace{-5pt}
    \caption{Late-layer zoom (layers 24--36) comparing the non-looped baseline, gated UpLoop, and ungated UpLoop (loop=3). Left: last-token hidden update RMS. Middle: cosine similarity between adjacent hidden states. Right: cosine similarity to the final decoder output. Gating suppresses late-layer write amplification, preserves better hidden-state continuity, and still allows gradual progress toward the final representation.}
    \label{fig:decoding}
    \vspace{-5pt}
\end{figure*}

We zoom in on decoder layers and compare gated and ungated loop-3 models against the non-looped baseline.
\Cref{fig:decoding} reports last-token hidden update RMS, adjacent hidden-state cosine, and cosine to the final decoder output.
First, ungated recurrence produces systematically larger late-layer writes: its hidden update RMS stays above both the gated model and the baseline, with the gap widening near the top of the decoder. Gating keeps these updates closer to the base scale.
Second, gating preserves local continuity: the ungated model shows a sharper drop in adjacent hidden-state cosine near the output-adjacent region, whereas the gated model remains closer.
Third, both looped models move progressively toward the final decoder output, but the gated does so more gradually.

\paragraph{Post-gate residual writes confirm that later loops operate in a refinement regime.}

We next measure residual writes \emph{after gating} to test whether later loops are controlled refinements rather than base-scale re-executions.
Specifically, we compare the residual write magnitudes at $r=2$ and $r=3$ with the base write.

\begin{figure}[t]
    \centering
    \includegraphics[width=0.56\linewidth]{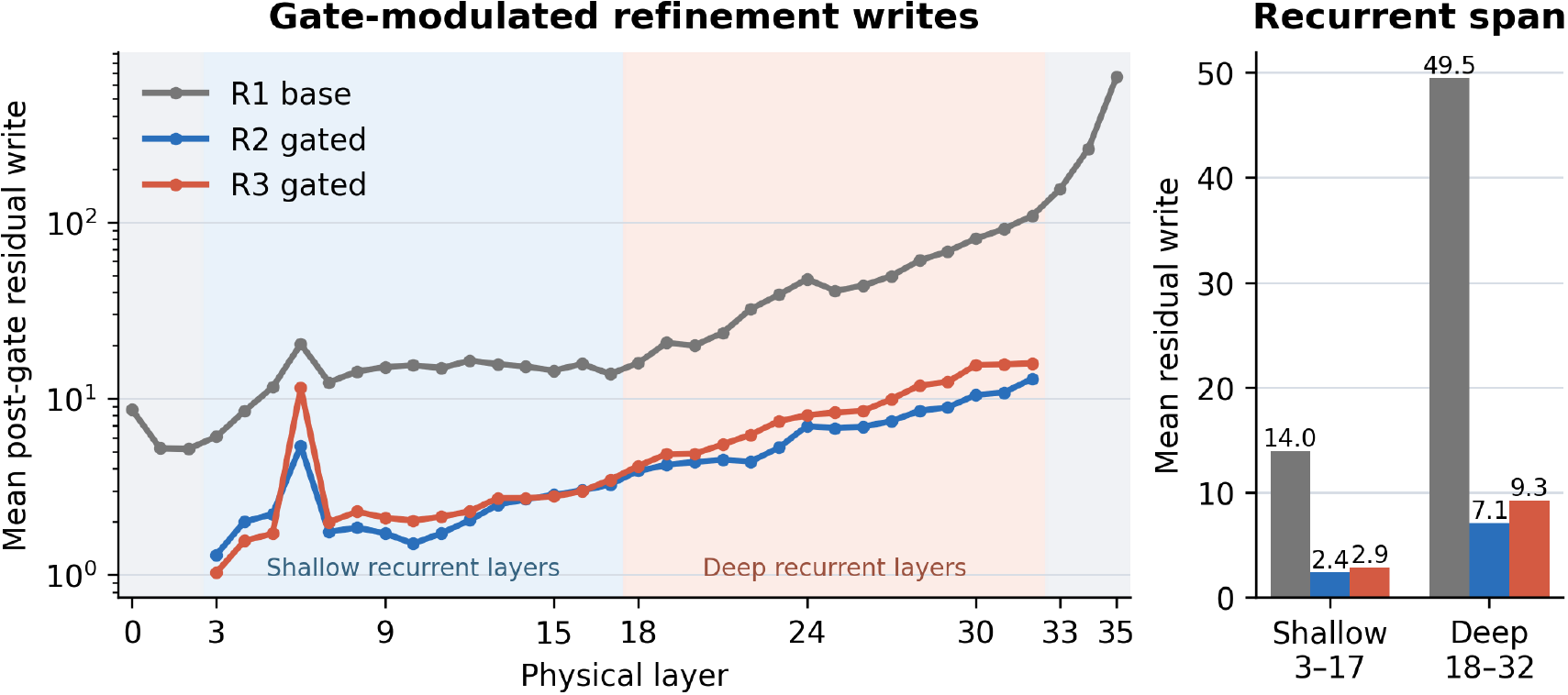}
    \caption{Gate-modulated residual write magnitude across looped layers. $r=1$ is the base write, while $r=2$ and $r=3$ are gated refinement writes. Later-round writes remain more modest than the base pass, showing that recurrent refinement operates through controlled residual corrections.}
    \label{fig:gate_modulated_writes}
    \vspace{-10pt}
\end{figure}

\Cref{fig:gate_modulated_writes} shows a consistent pattern.
Across shallow and deep spans, the gated writes at $r=2$ and $r=3$ are only a small fraction of the corresponding base write at $r=1$.
Quantitatively, they retain about 14\%--21\% of the $r=1$ magnitude.
So later loops do not rewrite the residual stream at the original scale; they apply modest corrections.
At the same time, $r=3$ is consistently larger than $r=2$, showing that the third loop remains active and often contributes more than the second refinement step, while still staying well below $r=1$.

\paragraph{Overall, gating turns repeated execution into controlled refinement.}

Taken together, these results show that gating makes recurrence better conditioned.
It suppresses excessive overwrite, preserves local continuity, and keeps later corrections meaningful but bounded.

\subsection{Learned Gates Allocate Refinement Non-Uniformly}
\label{sec:gate_profiles}

The previous subsection showed that gates make recurrent updates smaller and better behaved.
We next ask how this refinement capacity is allocated across depth, branches, and hidden dimensions.

\begin{figure*}[t]
    \centering
    \includegraphics[width=0.6\textwidth]{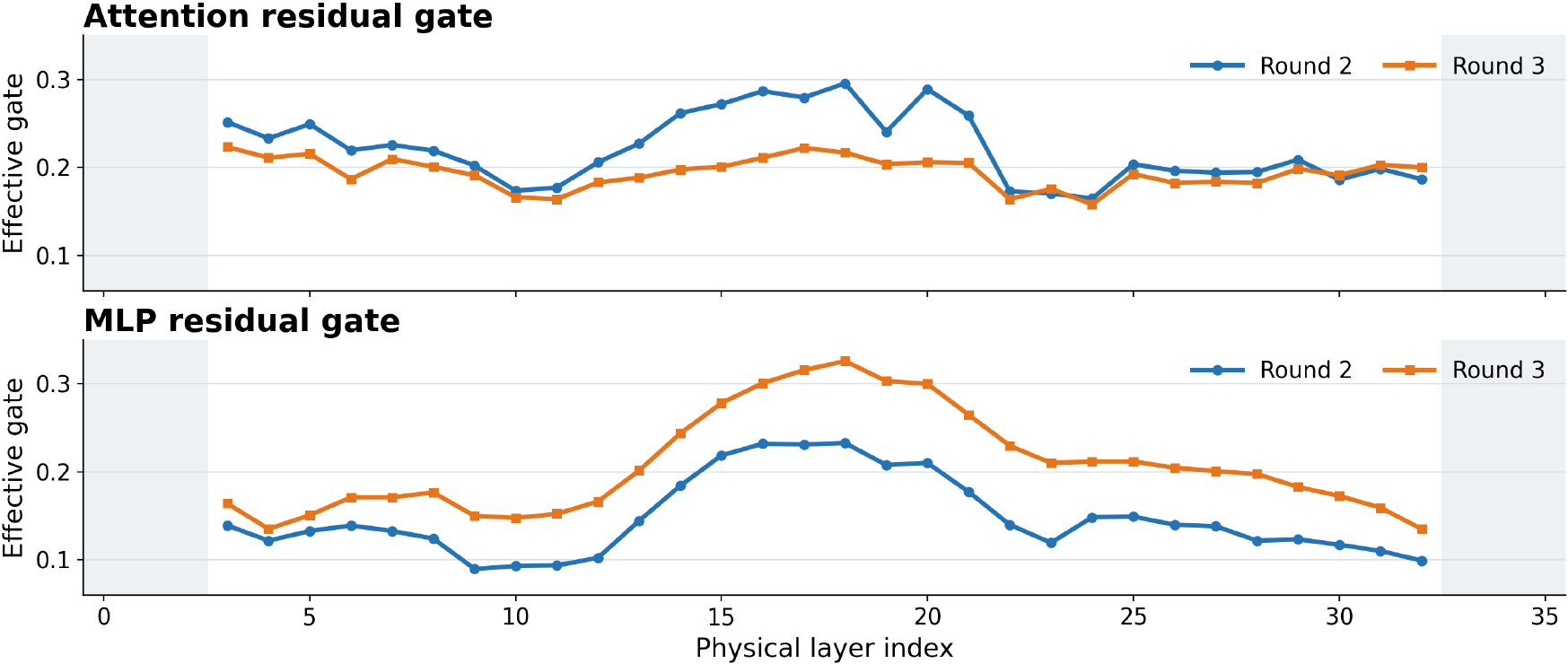}
    \caption{Learned static gate values for later loops in the $R=3$ model. The $r=2$ attention gates are often larger, while the $r=3$ MLP gates are often more prominent, suggesting that the second loop contributes relatively more to contextual re-access and the third loop relatively more to feature consolidation.}
    \label{fig:gate_profiles}
\end{figure*}

We first inspect the learned static gate values across physical depth.
\Cref{fig:gate_profiles} shows the mean effective gate values for the attention and MLP residual branches at rounds 2 and 3 in the loop-3 model.
The gate profiles are clearly non-uniform across depth, indicating that the model does not use looping uniformly across layers.
Instead, it allocates stronger looped writes to some layers and weaker ones to others.
In attention, the $r=2$ gate is often slightly larger than the $r=3$ gate, suggesting that the second loop contributes relatively more to re-accessing contextual information.
In MLP, the $r=3$ gate is often larger, especially in the middle layers, suggesting that the later loop contributes relatively more to feature transformation.

This loop-specific asymmetry suggests that later loops are not redundant repeats.
Rather, different loops appear to play different functional roles.
This is consistent with the analysis that gating places looping in a controlled small-step regime.

\begin{figure}[t]
    \centering
    \includegraphics[width=0.6\linewidth]{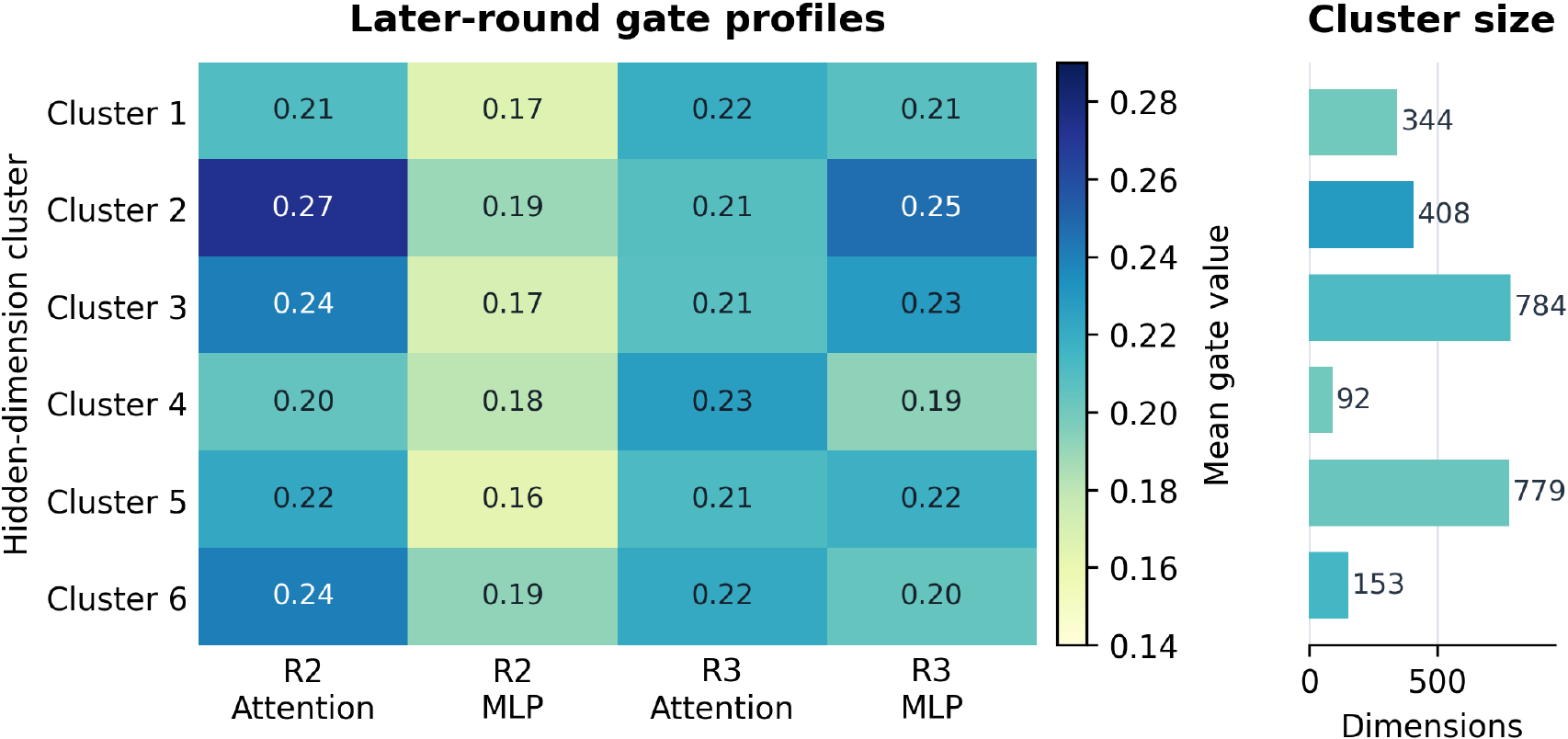}
    \caption{Hidden-dimension clusters of later-round gate profiles. Each hidden dimension is represented by its four-value profile $[\mathrm{r=2\ Attn}, \mathrm{r=2\ MLP}, \mathrm{r=3\ Attn}, \mathrm{r=3\ MLP}]$, standardized, and clustered with K-means ($k=6$). Different clusters exhibit distinct round- and branch-dependent patterns, indicating that recurrent refinement is distributed non-uniformly across hidden dimensions.}
    \label{fig:gate_subspace_clusters}
\end{figure}

To examine how recurrent refinement is allocated within hidden space, we further analyze the learned later-round gates at the hidden-dimension level.
For each hidden dimension, we form a four-value profile
\[
[\mathrm{r=2\ Attn},\ \mathrm{r=2\ MLP},\ \mathrm{r=3\ Attn},\ \mathrm{r=3\ MLP}],
\]
where each value is averaged across the looped physical layers.
We standardize these profiles and cluster hidden dimensions with deterministic K-means ($k=6$).
\Cref{fig:gate_subspace_clusters} shows that the learned gates are not uniform across hidden dimensions.
Instead, they form several distinct subspaces with different round- and branch-dependent profiles.
This indicates that later-round refinement is routed through differentiated hidden subspaces rather than controlled by a global gate strength.

Overall, the gate controls not only \emph{which layers} use more recurrent refinement, but also \emph{which hidden subspaces} are emphasized at different loops and branches.
This structured allocation helps explain why UpLoop remains controlled yet non-trivial: recurrence is neither uniformly suppressed nor uniformly amplified, but selectively deployed where additional refinement is most useful.

\subsection{Inference-Time Loop Extrapolation Beyond Training Depth}
\label{sec:loop_extrapolation}

A natural question for loop-depth models is whether useful computation can be \emph{extrapolated} beyond the loop count seen during training.
This question is especially relevant because UpLoop is obtained by \emph{SFT upcycling} from a strong pretrained non-looped LLM, rather than pretrained from scratch as a native looped model.

To test this, we take the model trained with maximum local loop count $R=3$ and perform \emph{inference-time loop extrapolation} to $R=4$ only at test time.
Concretely, the extra fourth loop reuses the same gated residual modulation as $r=3$, i.e., the $r=4$ attention and MLP gates are set equal to the learned $r=3$ gates.
This provides a simple test of whether the learned loop dynamics remain stable enough.

\begin{table*}[t]
\centering
\small
\resizebox{0.85\linewidth}{!}{
\begin{tabular}{llcccc}
\toprule
Category & Dataset & ($R=3$) & Extrapolated ($R=4$) & Native ($R=4$) & $\Delta$ (Extrapolated - Native) \\
\midrule
\multirow{5}{*}{Math}
& Math500           & \textbf{85.60} & 84.60 & 84.00 & +0.60 \\
& GSM-Plus          & 84.97 & 84.85 & \textbf{85.03} & -0.18 \\
& OlympiadBench     & \textbf{60.81} & 59.52 & 59.00 & +0.52 \\
& AIME 2025         & 36.61 & 34.66 & \textbf{39.53} & -4.87 \\
& AIME 2026         & 36.09 & 33.99 & \textbf{39.27} & -5.28 \\
\midrule
\multirow{6}{*}{Code}
& HumanEval+        & 68.29 & 71.95 & \textbf{73.78} & -1.83 \\
& MBPP+             & 60.32 & \textbf{64.81} & 64.29 & +0.52 \\
& LiveCodeBench v6  & 25.77 & \textbf{27.26} & 26.43 & +0.83 \\
& BigCodeBench      & 33.28 & \textbf{33.61} & 30.88 & +2.73 \\
& CruxEval          & 65.75 & \textbf{68.50} & 66.81 & +1.69 \\
& Aider             & 43.61 & \textbf{43.86} & 37.59 & +6.27 \\
\midrule
\multirow{4}{*}{Reasoning}
& MuSR              & 80.18 & 80.74 & \textbf{83.36} & -2.62 \\
& BBH               & 79.66 & \textbf{80.72} & 79.71 & +1.01 \\
& GPQA-Diamond      & 44.48 & 44.73 & \textbf{46.88} & -2.15 \\
& ZebraLogic        & 70.03 & \textbf{71.00} & 69.33 & +1.67 \\
\midrule
\multirow{5}{*}{General}
& BFCL-v3           & \textbf{66.14} & 65.91 & 63.21 & +2.70 \\
& PIQA              & 84.33 & 84.19 & \textbf{84.60} & -0.41 \\
& KorBench          & \textbf{56.56} & 56.36 & 57.52 & -1.16 \\
& IFBench           & 21.47 & 22.21 & \textbf{23.13} & -0.92 \\
& MMLU-Pro          & 57.54 & 57.61 & \textbf{59.56} & -1.95 \\
\midrule
\multicolumn{2}{l}{Average} & 58.07 & 58.55 & \textbf{58.70} & -0.15 \\
\bottomrule
\end{tabular}}
\caption{Inference-time loop extrapolation. The $r=4$ is not trained; it reuses the round-3 gates for both attention and MLP residual writes. The extrapolated loop improves over $R=3$ and nearly matches the performance of a natively trained $R=4$ model.}
\label{tab:loop_extrapolation}
\end{table*}

\Cref{tab:loop_extrapolation} shows that UpLoop supports useful loop extrapolation beyond its training depth. Extrapolating a model trained at $R=3$ to test-time $R=4$ improves the average from 58.07 to 58.55.
The gains, however, are highly domain-dependent. 
The clearest benefit appears in code generation and program understanding, including HumanEval+ (+3.66), MBPP+ (+4.49), LiveCodeBench v6 (+1.49), and CruxEval (+2.75). 
Several general or reasoning-oriented tasks also improve modestly. 
In contrast, mathematical benchmarks degrade consistently under $R=4$. 
This suggests that an additional loop is not uniformly beneficial.

Importantly, the extrapolated $R=4$ model is also close to the natively trained $R=4$ model: 58.55 versus 58.70 on average. This suggests that the learned recurrent dynamics are well structured and robust.

\subsection{Generalization to MoE Backbones}
\label{sec:moe_generalization}

To verify that UpLoop is not specific to dense Transformer backbones, we further test it on a \textbf{Mixture-of-Experts (MoE)} model, \textbf{Ling-3.0 Tiny}~\citep{ling3tiny_hf}. This model has about 7B parameters in total (1B activated parameters) with a hybrid dense--MoE design: the first layer uses a dense FFN, while the remaining 23 layers use routed MoE FFNs with 128 experts per layer and top-8 expert activation per token.

We evaluate the non-looped and UpLoop-$R=3$ variants on 17 benchmarks spanning math, code, tool use, reasoning, and knowledge-intensive tasks.
As shown in \Cref{tab:moe_generalization}, UpLoop transfers well to this MoE backbone: the average score improves from 50.97 to 55.51, yielding a gain of \textbf{+4.54} points. The improvements are broad, covering all 17 measured benchmarks. In particular, we observe consistent gains on math (Math500: +4.80, GSM8K-Plus: +7.25, OlympiadBench: +3.33), code (MBPP+: +5.87, CruxEval: +5.31, Aider: +9.51), reasoning (MuSR: +3.55, BBH: +5.15, KorBench: +7.36), and knowledge-heavy and general evaluation (MMLU-Pro: +6.28, PIQA: +6.10). This provides evidence that the proposed upcycling recipe generalizes beyond dense LLMs and can also serve as a practical depth scaling mechanism for MoE backbones.

\begin{table}[t]
\centering
\small
\resizebox{0.45\linewidth}{!}{
\begin{tabular}{lccc}
\toprule
Dataset & Non-looped & UpLoop ($R=3$) & $\Delta$ \\
\midrule
Math500          & 74.80 & \textbf{79.60} & +4.80 \\
GSM8K-Plus       & 70.77 & \textbf{78.02} & +7.25 \\
OlympiadBench    & 49.37 & \textbf{52.70} & +3.33 \\
HumanEval+       & 77.59 & \textbf{79.46} & +1.87 \\
MBPP+            & 66.68 & \textbf{72.55} & +5.87 \\
LiveCodeBench v6 & 23.51 & \textbf{25.61} & +2.10 \\
BigCodeBench     & 30.53 & \textbf{34.47} & +3.94 \\
CruxEval         & 51.31 & \textbf{56.62} & +5.31 \\
Aider            & 30.34 & \textbf{39.85} & +9.51 \\
BFCL Live        & 37.92 & \textbf{40.96} & +3.04 \\
PIQA             & 79.75 & \textbf{85.85} & +6.10 \\
MuSR             & 60.02 & \textbf{63.57} & +3.55 \\
BBH              & 61.71 & \textbf{66.86} & +5.15 \\
KorBench         & 39.84 & \textbf{47.20} & +7.36 \\
IFBench          & 20.13 & \textbf{21.20} & +1.07 \\
GPQA             & 45.01 & \textbf{45.71} & +0.70 \\
MMLU-Pro         & 47.13 & \textbf{53.41} & +6.28 \\
\midrule
Average          & 50.97 & \textbf{55.51} & \textbf{+4.54} \\
\bottomrule
\end{tabular}}
\caption{Generalization of UpLoop to an MoE backbone on 17 benchmarks. We evaluate Ling-3.0 Tiny, a hybrid dense--MoE model. UpLoop consistently improves over the non-looped baseline, showing that the proposed recurrent-depth upcycling recipe also generalizes to routed MoE architectures.}
\label{tab:moe_generalization}
\end{table}

\section{Training Cost Comparison}
\label{sec:training_cost}

\begin{figure*}[t]
    \centering
    \begin{subfigure}[t]{0.45\textwidth}
        \centering
        \includegraphics[width=\linewidth]{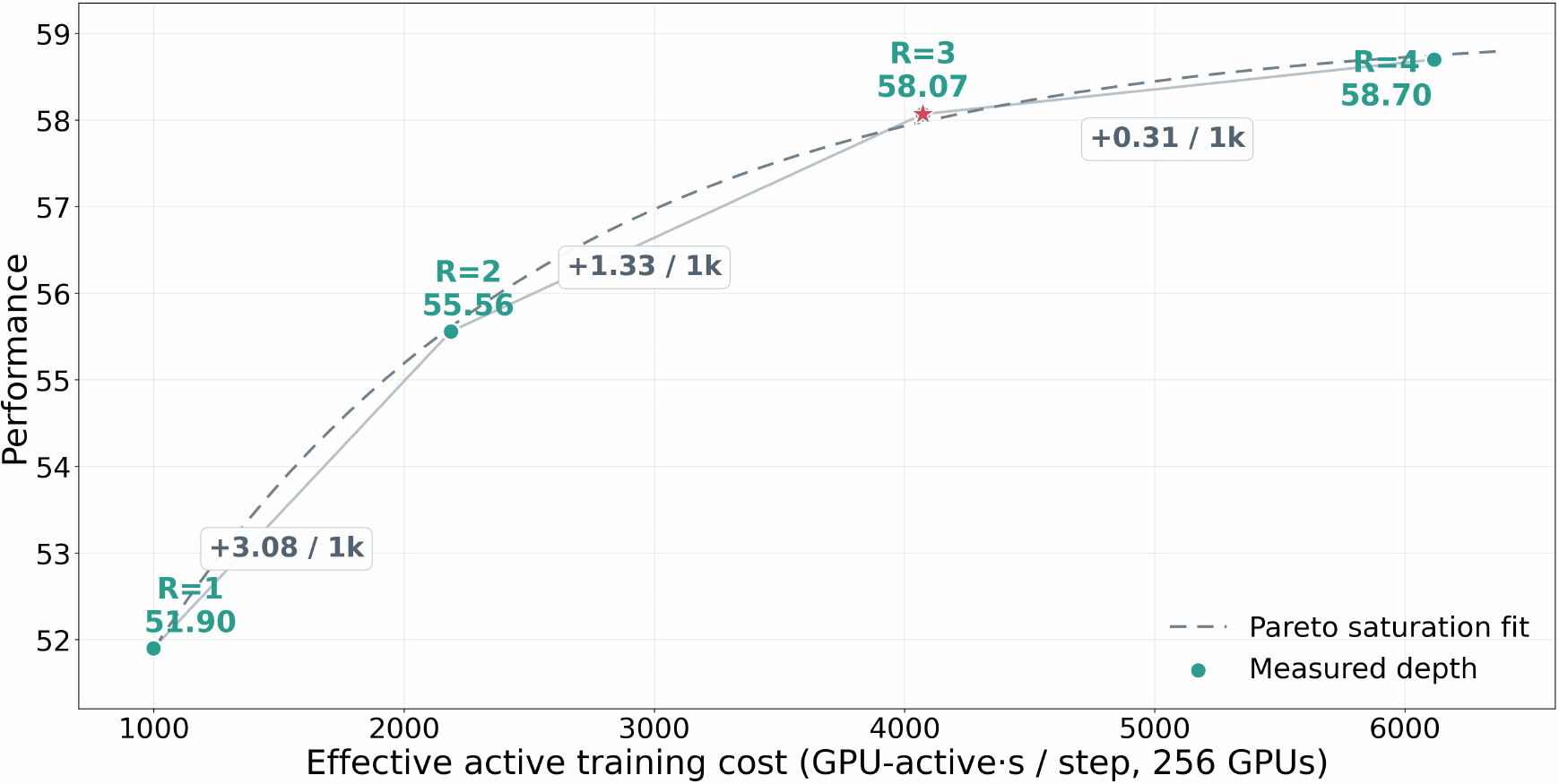}
        \caption{Performance vs. training cost.}
        \label{fig:training_cost_tradeoff}
    \end{subfigure}
    \hspace{0.05\textwidth}
    \begin{subfigure}[t]{0.45\textwidth}
        \centering
        \includegraphics[width=\linewidth]{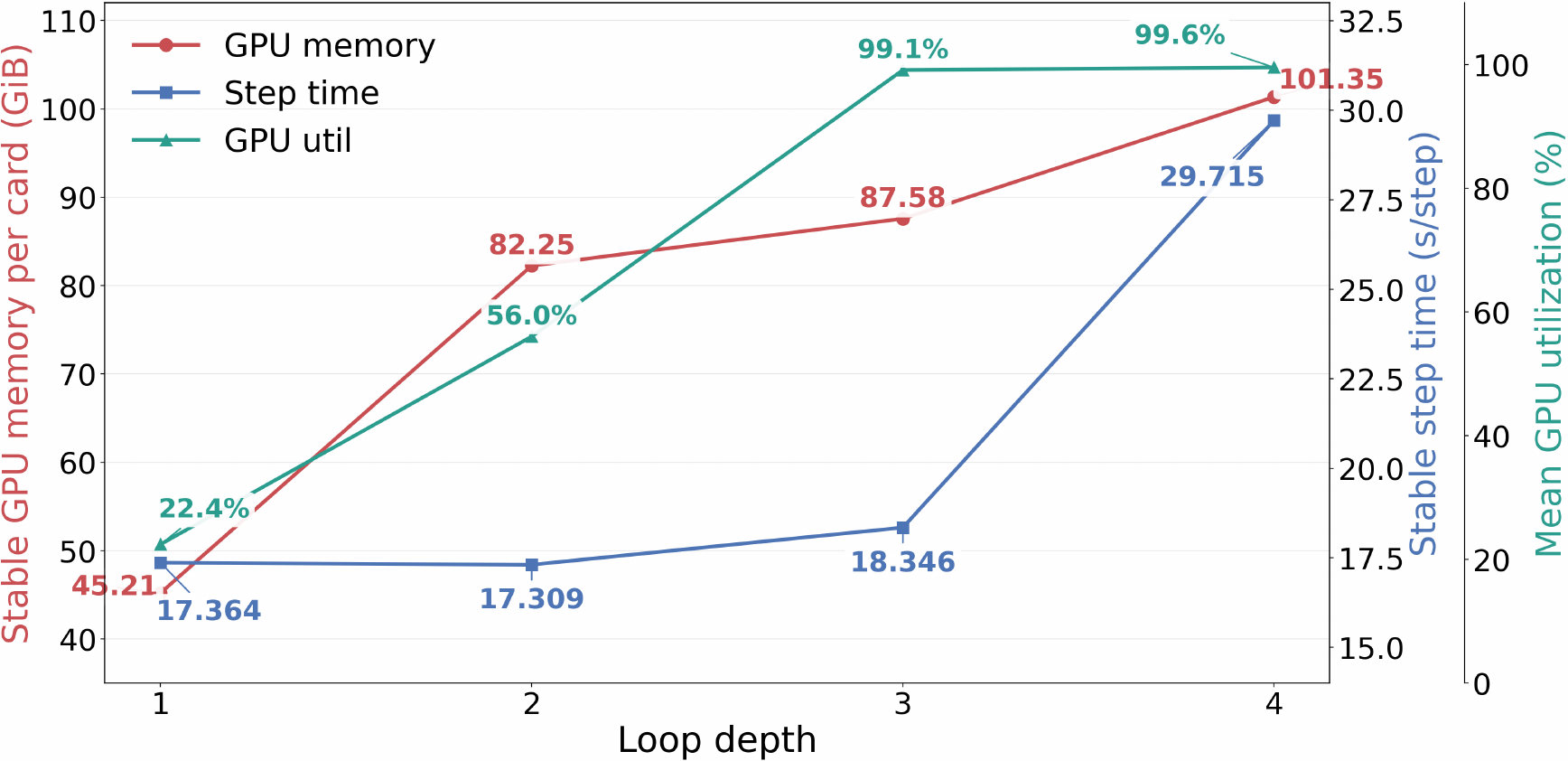}
        \caption{GPU memory and step time vs.  depth.}
        \label{fig:training_cost_memory_time}
    \end{subfigure}
    \vspace{-12pt}
\end{figure*}

We next assess whether the gains from increasing recurrent depth justify the added training cost. As shown in \Cref{fig:training_cost_tradeoff}, performance improves monotonically from 51.90 at $R=1$ to 55.56, 58.07, and 58.70 at $R=2,3,4$, respectively. However, the marginal gain diminishes sharply beyond $R=3$: increasing from $R=3$ to $R=4$ yields only +0.63 average points, making $R=3$ the most favorable accuracy--cost operating point.

\Cref{fig:training_cost_memory_time} further shows how the overhead evolves \emph{within the native $R=4$ training run} under the staged loop-depth curriculum, as training switches from the $r=1$ stage to $r=2$, $r=3$, and finally $r=4$. 
The resource demand rises progressively across stages: GPU memory increases from \mbox{$\sim$39--40\,GiB} at $r=1$ to \mbox{$\sim$70--75\,GiB}, \mbox{$\sim$73--81\,GiB}, and \mbox{$\sim$93--95\,GiB} at $r=2,3,4$, respectively. 
GPU utilization also climbs sharply from 22.5\% to 56.1\% and then 99.1\%, while iteration time stays nearly unchanged up to $r=3$ (17.36, 17.33, and 18.37\,s). 
This suggests that increasing depth up to $r=3$ mainly improves hardware utilization with little runtime penalty. By contrast, the transition from $r=3$ to $r=4$ yields only a marginal utilization gain (99.1\% $\rightarrow$ 99.6\%) but a sharp increase in iteration time and memory. Thus, beyond $r=3$, the additional loop contributes disproportionately more cost than efficiency.
Overall, UpLoop extends the useful loop-depth regime beyond the commonly observed $R=2$ ceiling, while $R=3$ remains the best practical trade-off.

\section{Inference Speed Optimization}

\subsection{Scheduling-Level Optimization: Self-Speculative Decoding}
\label{sec:self_spec}

\paragraph{Background and Method.}
While looped models improve representation through repeated computation, executing multiple loops inevitably increases decoding latency. 
Recently, self-speculative decoding for block- or model-wise looped Transformers, represented by LoopSpec~\citep{cho2026loopspec}, has attracted significant attention by directly extracting draft tokens from early recurrent states without an auxiliary draft model and verifying them against the full-depth looped target for lossless acceleration.
Inspired by this paradigm, we adapt the self-speculative principle to accelerate our \emph{layer-wise} looped architecture, UpLoop.

Because UpLoop is trained with intermediate supervision (\Cref{sec:method_objective}), the single-pass execution ($r=1$) remains directly decodable by the language modeling head, while its top-1 predictions already agree with the final $R=3$ output on more than 96\% of positions (\Cref{sec:progressive_refinement}). This makes the $r=1$ pass a natural draft model embedded within the target network itself. 
Concretely, for our main $R=3$ UpLoop model, we employ an \textbf{$r=1$ drafting and full $R=3$ verification} pipeline without loading any external draft model:
\begin{enumerate}[leftmargin=1.5em]
    \item \textbf{Drafting via $r=1$:} Starting from the committed prefix, the model executes only the first local loop ($r=1$) at each layer to autoregressively propose up to $K=6$ candidate tokens. To prevent low-quality rollouts, we apply confidence-based early stopping: if the top-1 probability of a drafted token falls below a threshold of $\tau=0.15$, drafting terminates immediately.
    \item \textbf{Parallel Verification via $R=3$:} The model then performs a single batched forward pass using the full $R=3$ recurrent computation ($r=1,2,3$) over all drafted candidate tokens simultaneously. Candidate tokens are verified sequentially against the target model's greedy predictions; at the first mismatch, the divergent draft token is replaced by the target token from $R=3$, and the next drafting cycle begins.
\end{enumerate}
Under greedy decoding, this verification rule guarantees that the generated sequence is \textbf{mathematically identical} to standard $R=3$ autoregressive decoding, as it merely amortizes memory-bound step-by-step $R=3$ calls into compute-efficient parallel verification passes.

\paragraph{Experimental Setup and Fair Comparison.}
To rigorously evaluate the wall-clock acceleration across different deployment scenarios, we benchmark self-speculative decoding on 14 representative tasks under two mainstream inference backends: \textbf{HuggingFace Transformers} (low-concurrency / local execution) and \textbf{SGLang} (high-concurrency serving with tensor parallelism $\mathrm{TP}=8$, \texttt{max\_running\_requests=1024}, and \texttt{mem\_fraction\_static=0.80}). 
To ensure a strictly fair comparison and isolate the algorithmic gain of self-speculation from KV-cache approximations, both standard $R=3$ decoding and self-speculative decoding use greedy decoding. 
Across all runs, we measure throughput in tokens per second (TPS) by dividing the exact number of generated completion tokens by the end-to-end wall-clock batch execution time.

\begin{table*}[t]
\centering
\small
\resizebox{0.92\linewidth}{!}{
\begin{tabular}{llcccccc}
\toprule
& & \multicolumn{3}{c}{HuggingFace Backend} & \multicolumn{3}{c}{SGLang Backend ($\mathrm{TP}=8$)} \\
\cmidrule(lr){3-5} \cmidrule(lr){6-8}
Category & Dataset & Standard TPS & Self-Spec TPS & Speedup & Standard TPS & Self-Spec TPS & Speedup \\
\midrule
\multirow{3}{*}{Math}
& Math500           & 88.60 & \textbf{139.06} & \textbf{1.57$\times$} & 1386.63 & \textbf{2816.76} & \textbf{2.03$\times$} \\
& GSM-Plus          & 89.99 & \textbf{184.32} & \textbf{2.05$\times$} & 617.11  & \textbf{1215.95} & \textbf{1.97$\times$} \\
& AIME 2025         & 62.56 & \textbf{117.24} & \textbf{1.87$\times$} & 2429.75 & \textbf{3695.65} & \textbf{1.52$\times$} \\
\midrule
\multirow{4}{*}{Code}
& MBPP+             & 48.82 & \textbf{80.96}  & \textbf{1.66$\times$} & 346.44  & \textbf{678.22}  & \textbf{1.96$\times$} \\
& LiveCodeBench v6  & 54.88 & \textbf{130.28} & \textbf{2.37$\times$} & 2090.50 & \textbf{4244.76} & \textbf{1.61$\times$} \\
& BigCodeBench      & 84.50 & \textbf{163.36} & \textbf{1.93$\times$} & 614.56  & \textbf{1473.30} & \textbf{2.40$\times$} \\
& CruxEval          & 66.75 & \textbf{132.88} & \textbf{1.99$\times$} & 1213.82 & \textbf{2371.83} & \textbf{1.95$\times$} \\
\midrule
\multirow{3}{*}{Reasoning}
& MuSR              & 78.18 & \textbf{117.79} & \textbf{1.51$\times$} & 941.50  & \textbf{1903.04} & \textbf{2.02$\times$} \\
& BBH               & 13.83 & \textbf{25.88}  & \textbf{1.87$\times$} & 1485.85 & \textbf{1995.17} & \textbf{1.99$\times$} \\
& GPQA-Diamond      & 76.51 & \textbf{122.77} & \textbf{1.60$\times$} & 2346.14 & \textbf{3672.77} & \textbf{1.57$\times$} \\
\midrule
\multirow{4}{*}{General}
& PIQA              & 67.97 & \textbf{115.61} & \textbf{1.70$\times$} & 384.14  & \textbf{994.26}  & \textbf{2.59$\times$} \\
& KorBench          & 28.28 & \textbf{47.99}  & \textbf{1.70$\times$} & 2005.31 & \textbf{3465.18} & \textbf{1.73$\times$} \\
& IFBench           & 55.16 & \textbf{105.39} & \textbf{1.91$\times$} & 2147.85 & \textbf{3415.91} & \textbf{1.59$\times$} \\
& MMLU-Pro          & 41.58 & \textbf{78.68}  & \textbf{1.89$\times$} & 1571.18 & \textbf{2700.57} & \textbf{1.72$\times$} \\
\midrule
\multicolumn{2}{l}{Average (14)} & 61.26 & \textbf{111.59} & \textbf{1.83$\times$} & 1398.63 & \textbf{2474.53} & \textbf{1.90$\times$} \\
\bottomrule
\end{tabular}}
\caption{Throughput (tokens/s, TPS) and wall-clock speedup of self-speculative decoding ($r=1$ drafting, full $R=3$ verification) for Qwen3-4B UpLoop ($R=3$) across 14 benchmarks under both HuggingFace and SGLang ($\mathrm{TP}=8$) backends. All runs use standard independent KV caches and lossless greedy decoding.}
\label{tab:self_spec_main}
\end{table*}

\paragraph{Results and Analysis.}
\Cref{tab:self_spec_main} summarizes the throughput and speedup results across all 14 benchmarks.
\begin{itemize}[leftmargin=1.5em]
    \item \textbf{Consistent acceleration across tasks and serving engines:} Self-speculative decoding improves throughput on every evaluated benchmark. On HuggingFace, the average TPS increases from 61.26 to 111.59, achieving an average speedup of \textbf{1.83$\times$} (ranging from 1.51$\times$ on MuSR to 2.37$\times$ on LiveCodeBench v6). Under high-concurrency serving in SGLang, the gains remain equally pronounced: average TPS rises from 1398.63 to 2474.53, yielding an average speedup of \textbf{1.90$\times$} (reaching up to \textbf{2.59$\times$} on PIQA).
    \item \textbf{High draft acceptance from progressive refinement:} The strong speedups directly reflect the alignment between early and late loops. Because $r=1$ captures the majority of token-level decisions while $r=2,3$ focus on targeted local refinements, multi-token drafts proposed by $r=1$ achieve high acceptance rates.
    \item \textbf{Hardware-friendly execution:} Unlike conventional speculative decoding that requires hosting a separate draft model, our self-speculative scheme reuses the existing $r=1$ weights. In our HuggingFace profiling, this approach incurs only a negligible $+5.45\%$ average change in GPU memory while reducing average GPU utilization by $3.14\%$, confirming that the speedup comes from converting sequential memory-bound decoding steps into batched verification without introducing heavy resource contention.
\end{itemize}

\subsection{Architectural-Level Optimization: Cross-Loop KV Reuse}
\label{sec:kvreuse}

While self-speculative decoding (\Cref{sec:self_spec}) accelerates inference at the \emph{scheduling} level, each verification step still maintains and attends over three separate sets of key--value (KV) caches across local rounds ($r=1,2,3$). In long-context or high-concurrency serving, this increases both KV memory footprint and memory-bandwidth pressure. 
This raises a complementary \emph{architectural} question: Can we further reduce inference overhead by reusing KV caches across loops without materially hurting accuracy?

\paragraph{Accuracy and Cross-Loop Alignment of KV Reuse.}
Similar to LoopCoder-v2~\citep{yang2026loopcoder}, we first evaluate two training-free KV-reuse approximations in the $R=3$ setting:
(i) \textbf{KV-share$(3\leftarrow2)$}, where $r=3$ reuses the KV cache from $r=2$ while retaining its own query, output projection, and MLP computation; and
(ii) \textbf{KV-share$(2,3\leftarrow1)$}, where both $r=2$ and $r=3$ reuse the KV cache from $r=1$.

\begin{table*}[t]
\centering
\small
\resizebox{0.85\linewidth}{!}{
\begin{tabular}{llcccc}
\toprule
Category & Dataset & loop ($R=3$) & KV-share$(3\leftarrow2)$ & KV-share$(2,3\leftarrow1)$ & $\Delta$ share$(3\leftarrow2)$ \\
\midrule
\multirow{5}{*}{Math}
& Math500           & \textbf{85.60} & 84.40 & 83.00 & -1.20 \\
& GSM-Plus          & \textbf{84.97} & 84.74 & 83.72 & -0.23 \\
& OlympiadBench     & \textbf{60.81} & 60.41 & 58.76 & -0.40 \\
& AIME 2025         & 36.61 & \textbf{36.82} & 33.23 & +0.21 \\
& AIME 2026         & \textbf{36.09} & 34.84 & 32.86 & -1.25 \\
\midrule
\multirow{6}{*}{Code}
& HumanEval+        & \textbf{68.29} & 67.07 & 65.24 & -1.22 \\
& MBPP+             & 60.32 & \textbf{62.96} & 59.26 & +2.64 \\
& LiveCodeBench v6  & 25.77 & \textbf{26.93} & 25.00 & +1.16 \\
& BigCodeBench      & \textbf{33.28} & 31.70 & 30.88 & -1.58 \\
& CruxEval          & 65.75 & \textbf{65.94} & 62.94 & +0.19 \\
& Aider             & \textbf{43.61} & 42.11 & 35.83 & -1.50 \\
\midrule
\multirow{4}{*}{Reasoning}
& MuSR              & 80.18 & \textbf{81.39} & 79.39 & +1.21 \\
& BBH               & \textbf{79.66} & 79.32 & 75.59 & -0.34 \\
& GPQA-Diamond      & 44.48 & \textbf{45.61} & 42.55 & +1.13 \\
& ZebraLogic        & 70.03 & \textbf{70.53} & 60.30 & +0.50 \\
\midrule
\multirow{5}{*}{General}
& BFCL-v3           & \textbf{66.14} & 65.09 & 63.15 & -1.05 \\
& PIQA              & 84.33 & \textbf{84.35} & 83.21 & +0.02 \\
& KorBench          & \textbf{56.56} & 55.92 & 54.48 & -0.64 \\
& IFBench           & 21.47 & 22.80 & \textbf{24.93} & +1.33 \\
& MMLU-Pro          & 57.54 & \textbf{57.94} & 55.61 & +0.40 \\
\midrule
\multicolumn{2}{l}{Average} & \textbf{58.07} & 58.04 & 55.50 & -0.03 \\
\bottomrule
\end{tabular}}
\caption{KV-reuse evaluation. We compare the standard $R=3$ UpLoop model against two training-free approximations: reusing the $r=2$ KV cache at $r=3$, and reusing the $r=1$ KV cache at both $r=2$ and $r=3$. The last column reports the difference between standard $R=3$ and the $r=3 \leftarrow r=2$ KV-reuse variant.}
\label{tab:kvreuse_main}
\end{table*}

\begin{table}[t]
\centering
\small
\resizebox{0.75\linewidth}{!}{
\begin{tabular}{lccccccccc}
\toprule
Loop pair & Q & K & V & PV & Attn out & Top-1 & Top-8 & Mass & $X$ \\
\midrule
$1 \rightarrow 2$ & 0.9566 & 0.9656 & 0.8799 & 0.8507 & 0.7637 & 0.8597 & 0.8126 & 0.8404 & 0.9373 \\
$2 \rightarrow 3$ & \textbf{0.9882} & \textbf{0.9907} & \textbf{0.9758} & \textbf{0.9616} & \textbf{0.8948} & \textbf{0.9465} & \textbf{0.9059} & \textbf{0.9308} & \textbf{0.9853} \\
\bottomrule
\end{tabular}}
\caption{Similarity and overlap statistics between adjacent loops. ``PV'' denotes projected value similarity, ``Mass'' denotes attention-mass overlap, and $X$ denotes hidden-state cosine similarity.}
\label{tab:qkv}
\end{table}

\Cref{tab:kvreuse_main} reveals a clear asymmetry.
Reusing the $r=2$ KV cache at $r=3$ leaves overall accuracy essentially unchanged: the 20-benchmark mean moves from 58.07 to 58.04, indicating that $r=3$ can safely share the $r=2$ KV cache without performance degradation.
By contrast, reusing the $r=1$ KV cache for both later loops causes a noticeable drop to 55.50. 
To explain this asymmetry, we compare adjacent loops directly in representation and attention space.
As shown in \Cref{tab:qkv}, all similarity and overlap metrics are substantially higher for $r=2 \rightarrow r=3$ than for $r=1 \rightarrow r=2$, including Q/K/V cosine similarity ($>0.975$).
This confirms that the recurrent trajectory stabilizes at later loops: by $r=3$, the model largely agrees with $r=2$ on \emph{where} to attend, while still refining \emph{what} to write through its query and MLP branches. Forcing $r=2$ and $r=3$ to reuse the unrefined $r=1$ KV cache, however, removes essential contextual updates from the second loop.

\paragraph{Combining Self-Speculative Decoding with KV Sharing.}
Given that KV-share$(3\leftarrow2)$ is virtually lossless in accuracy, we next integrate it into our SGLang ($\mathrm{TP}=8$) self-speculative decoding pipeline (\Cref{sec:self_spec}). Under this combined setting, $r=1$ drafts candidate tokens as before, while during the parallel $R=3$ verification pass, $r=3$ directly reuses the K/V projections and cached states computed at $r=2$.

\begin{table*}[t]
\centering
\small
\resizebox{0.98\linewidth}{!}{
\begin{tabular}{llccccc}
\toprule
Category & Dataset & Standard TPS & Self-Spec TPS & Self-Spec Speedup & \makecell{Self-Spec +\\KV-Share$(3\leftarrow2)$ TPS} & \makecell{Combined Speedup\\(vs.\ Standard)} \\
\midrule
\multirow{3}{*}{Math}
& Math500           & 1386.63 & 2816.76 & 2.03$\times$ & \textbf{3069.11} & \textbf{2.21$\times$} \\
& GSM-Plus          & 617.11  & 1215.95 & 1.97$\times$ & \textbf{1240.21} & \textbf{2.01$\times$} \\
& AIME 2025         & 2429.75 & 3695.65 & 1.52$\times$ & \textbf{3896.41} & \textbf{1.60$\times$} \\
\midrule
\multirow{4}{*}{Code}
& MBPP+             & 346.44  & 678.22  & 1.96$\times$ & \textbf{773.35}  & \textbf{2.23$\times$} \\
& LiveCodeBench v6  & 2090.50 & \textbf{4244.76} & 1.61$\times$ & 3972.64 & \textbf{1.90$\times$} \\
& BigCodeBench      & 614.56  & 1473.30 & 2.40$\times$ & \textbf{1544.89} & \textbf{2.51$\times$} \\
& CruxEval          & 1213.82 & 2371.83 & 1.95$\times$ & \textbf{2430.73} & \textbf{2.00$\times$} \\
\midrule
\multirow{3}{*}{Reasoning}
& MuSR              & 941.50  & 1903.04 & 2.02$\times$ & \textbf{2202.57} & \textbf{2.34$\times$} \\
& BBH               & 1485.85 & 1995.17 & 1.99$\times$ & \textbf{3401.30} & \textbf{2.29$\times$} \\
& GPQA-Diamond      & 2346.14 & 3672.77 & 1.57$\times$ & \textbf{3805.41} & \textbf{1.62$\times$} \\
\midrule
\multirow{4}{*}{General}
& PIQA              & 384.14  & \textbf{994.26}  & \textbf{2.59$\times$} & 967.81  & 2.52$\times$ \\
& KorBench          & 2005.31 & 3465.18 & 1.73$\times$ & \textbf{3543.42} & \textbf{1.77$\times$} \\
& IFBench           & 2147.85 & 3415.91 & 1.59$\times$ & \textbf{3914.94} & \textbf{1.82$\times$} \\
& MMLU-Pro          & 1571.18 & 2700.57 & 1.72$\times$ & \textbf{2983.71} & \textbf{1.90$\times$} \\
\midrule
\multicolumn{2}{l}{Average} & 1398.63 & 2474.53 & 1.90$\times$ & \textbf{2696.18} & \textbf{2.05$\times$} \\
\bottomrule
\end{tabular}}
\caption{Orthogonal speedup from combining self-speculative decoding with cross-loop KV sharing for Qwen3-4B UpLoop ($R=3$) in SGLang. Enabling KV-share$(3\leftarrow2)$ on top of self-speculation further boosts average throughput from 2474.53 to 2696.18 TPS, achieving an overall \textbf{2.05$\times$} wall-clock speedup.}
\label{tab:sglang_kv_combined}
\end{table*}

As shown in \Cref{tab:sglang_kv_combined}, self-speculative decoding and cross-loop KV reuse provide \textbf{orthogonal, stackable gains}. Under SGLang ($\mathrm{TP}=8$), enabling KV-share$(3\leftarrow2)$ on top of self-speculative decoding further lifts the average throughput from 2474.53 to \textbf{2696.18 TPS}, expanding the overall end-to-end speedup from $1.90\times$ to \textbf{2.05$\times$} (with peak speedups reaching \textbf{2.52$\times$} on PIQA, \textbf{2.51$\times$} on BigCodeBench, \textbf{2.34$\times$} on MuSR, and \textbf{2.29$\times$} on BBH). 
Overall, combining $r=1$ self-speculative drafting with $r=2/r=3$ KV cache sharing allows our $R=3$ UpLoop model to recover more than $2\times$ inference speed while simultaneously preserving accuracy.

\subsection{Scheduling-Level and Architectural-Level Optimization for UpLoop-Ling-3.0 Tiny}
\label{sec:moe_routing_reuse}

We further extend our inference optimization pipeline to the upcycled MoE backbone, \textbf{Ling-3.0 Tiny} ($R=3$). 
Compared with dense Transformers with standard full attention, Ling-3.0 Tiny relies on hybrid linear and compressed attention mechanisms (KDA~\citep{kda} and MLA~\citep{MLA}), so KV-cache is no longer the dominant bottleneck. Instead, the main overhead comes from repeatedly executing the MoE router, activating different experts, and maintaining loop-specific recurrent states in linear-attention layers.

\paragraph{Scheduling-Level Optimization: Self-Speculative Decoding on Ling-3.0 Tiny.}
We first evaluate the direct application of our self-speculative decoding scheme to the MoE backbone, using $r=1$ for drafting and full $R=3$ for verification. 
All serving experiments in this subsection are conducted with the SGLang runtime under a unified deployment configuration, with tensor parallelism $\mathrm{TP}=8$, \texttt{max\_running\_requests=1024}, \texttt{mamba\_cache=1024}, and \texttt{mem\_fraction\_static=0.68}. 
\Cref{tab:ling_moe_spec_tps} compares standard non-speculative decoding against self-speculative decoding on 12 representative benchmarks. 
Even without any additional cross-loop sharing, self-speculative decoding consistently improves throughput on every benchmark, increasing the average throughput from \textbf{1495.19} to \textbf{2546.96} tokens/s (\textbf{1.70$\times$} speedup).

\begin{table*}[t]
\centering
\small
\resizebox{0.6\linewidth}{!}{
\begin{tabular}{llccc}
\toprule
Category & Dataset & Non-Spec TPS & Self-Spec TPS & Speedup \\
\midrule
\multirow{1}{*}{Math}
& GSM-Plus          & 1067.67 & \textbf{1765.59} & \textbf{1.65$\times$} \\
\midrule
\multirow{5}{*}{Code}
& HumanEval+        & 1283.76 & \textbf{2141.91} & \textbf{1.67$\times$} \\
& MBPP+             & 1200.95 & \textbf{2246.60} & \textbf{1.87$\times$} \\
& LiveCodeBench v6  & 1812.62 & \textbf{2889.50} & \textbf{1.59$\times$} \\
& BigCodeBench      & 1280.31 & \textbf{2127.07} & \textbf{1.66$\times$} \\
& CruxEval          & 1569.55 & \textbf{2667.58} & \textbf{1.70$\times$} \\
\midrule
\multirow{2}{*}{Reasoning}
& BBH               & 1814.41 & \textbf{3084.03} & \textbf{1.70$\times$} \\
& GPQA-Diamond      & 1794.22 & \textbf{3119.96} & \textbf{1.74$\times$} \\
\midrule
\multirow{4}{*}{General}
& PIQA              & 747.47  & \textbf{1150.41} & \textbf{1.54$\times$} \\
& KorBench          & 1567.46 & \textbf{2731.22} & \textbf{1.74$\times$} \\
& IFBench           & 1902.79 & \textbf{3404.26} & \textbf{1.79$\times$} \\
& MMLU-Pro          & 1901.02 & \textbf{3235.44} & \textbf{1.70$\times$} \\
\midrule
\multicolumn{2}{l}{Average} & 1495.19 & \textbf{2546.96} & \textbf{1.70$\times$} \\
\bottomrule
\end{tabular}}
\caption{Throughput comparison (tokens/s, TPS) between standard decoding and self-speculative decoding on Ling-3.0 Tiny UpLoop ($R=3$). Self-speculative decoding alone improves average throughput by \textbf{1.70$\times$}.}
\label{tab:ling_moe_spec_tps}
\end{table*}

\begin{figure}[t]
    \centering
    \includegraphics[width=0.82\linewidth]{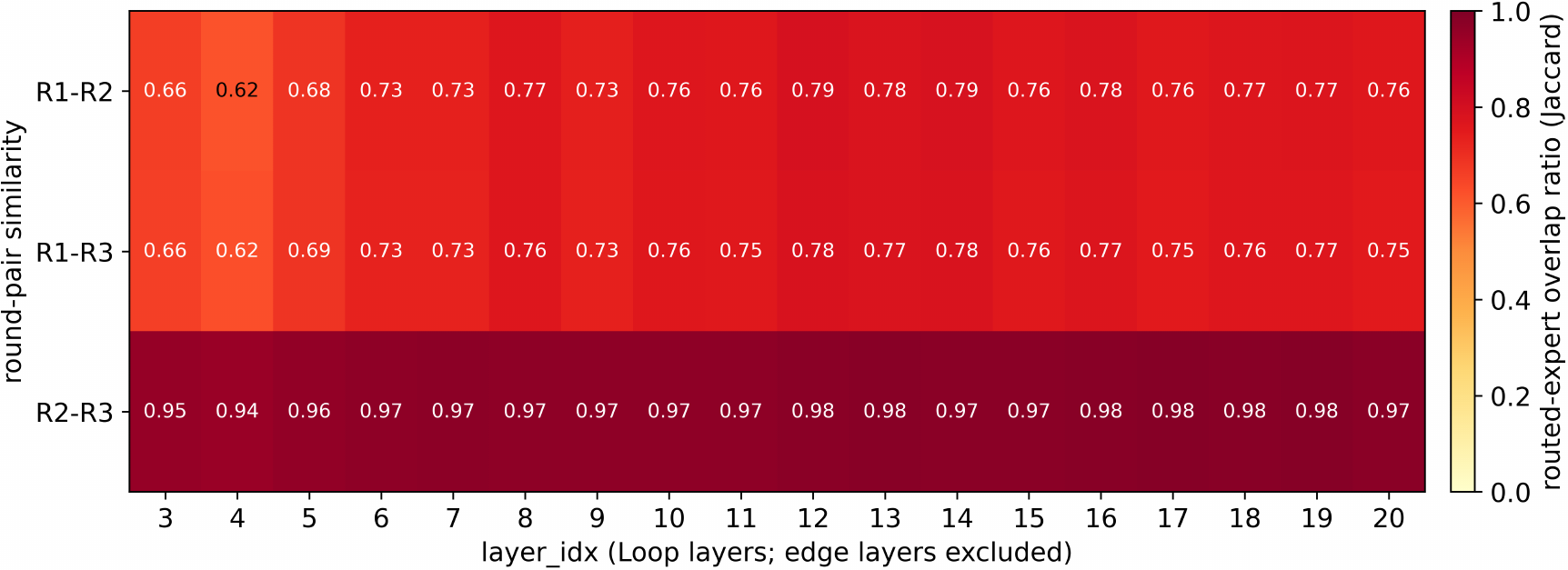}
    \caption{Pairwise routed-expert overlap (Jaccard similarity) across looped layers (layers 3--20) in Ling-3.0 Tiny UpLoop ($R=3$). While $r=1 \rightarrow r=2$ and $r=1 \rightarrow r=3$ exhibit moderate overlap, the activated top-8 experts between $r=2$ and $r=3$ are highly consistent across all depths.}
    \label{fig:ling_routing_overlap}
\end{figure}

\paragraph{Architectural-Level Optimization for UpLoop-Ling-3.0 Tiny.}
The strong gains from self-speculative decoding suggest that loop redundancy can also be exploited in MoE backbones. 
However, unlike UpLoop-Qwen3-4B with GQA~\citep{gqa} where the main opportunity lies in reducing repeated attention/KV work, UpLoop-Ling-3.0 Tiny exposes two additional forms of cross-loop redundancy.
First, the router may already stabilize on nearly identical expert choices after the second local loop. 
Indeed, as shown in \Cref{fig:ling_routing_overlap}, the top-8 routed experts between $r=2$ and $r=3$ exhibit very high overlap across all looped middle layers, while $r=1 \rightarrow r=2$ and $r=1 \rightarrow r=3$ remain noticeably less aligned. 
This suggests that the router has largely converged by $r=2$, making the third-loop routing computation potentially redundant.
Second, Ling-3.0 Tiny contains KDA-based linear-attention layers with recurrent \texttt{conv}/SSM state. 
In the default implementation, $r=2$ and $r=3$ maintain separate recurrent-state streams for the same physical KDA layer. 
We find that for verification-only execution, these streams can also be reused safely across local loops: specifically, the post-$r=2$ recurrent state of a physical KDA layer can be handed off directly to $r=3$, which then updates the same stream in place. 
This eliminates the need to maintain an additional independent recurrent-state slot for $r=3$ while reducing recurrent-state management overhead during verification.

\paragraph{Two Training-Free Cross-Loop Reuse Strategies.}
Based on these observations, we consider two architectural-level training-free optimizations on top of self-speculative decoding:

\begin{itemize}
    \item \textbf{Routing-Share$(3\leftarrow2)$}: during the third local loop ($r=3$), we skip router gate computation and top-$k$ selection, and directly reuse the expert indices and routing weights selected at $r=2$.
    \item \textbf{Recurrent-Share$(3\leftarrow2)$}: for each KDA layer, we let $r=3$ reuse the same recurrent-state stream updated by $r=2$, instead of allocating and maintaining a separate loop-specific recurrent-state slot.
\end{itemize}

\begin{table*}[t]
\centering
\small
\resizebox{0.79\linewidth}{!}{
\begin{tabular}{llccc}
\toprule
Category & Dataset & \makecell{Self-Spec Only\\Acc. / TPS} & \makecell{+Routing-Share$(3\leftarrow2)$\\Acc. / TPS} & \makecell{+Recurrent-Share$(3\leftarrow2)$\\Acc. / TPS} \\
\midrule
\multirow{1}{*}{Math}
& GSM-Plus
& 78.02 / 1765.59
& 78.12 / 1754.15
& 77.14 / 2289.08 \\
\midrule
\multirow{5}{*}{Code}
& HumanEval+
& 79.46 / 2141.91
& 78.19 / 2201.84
& 78.63 / 2686.26 \\
& MBPP+
& 72.55 / 2246.60
& 72.78 / 2163.00
& 73.31 / 2889.40 \\
& LiveCodeBench v6
& 25.61 / 2889.50
& 25.39 / 3179.35
& 26.05 / 3814.23 \\
& BigCodeBench
& 34.47 / 2127.07
& 33.33 / 2121.16
& 32.89 / 2894.28 \\
& CruxEval
& 56.62 / 2667.58
& 56.38 / 2829.86
& 55.44 / 3804.19 \\
\midrule
\multirow{2}{*}{Reasoning}
& BBH
& 66.86 / 3084.03
& 67.17 / 3109.14
& 66.37 / 4319.65 \\
& GPQA-Diamond
& 45.71 / 3119.96
& 45.83 / 3124.05
& 45.17 / 4243.32 \\
\midrule
\multirow{4}{*}{General}
& PIQA
& 85.85 / 1150.41
& 86.18 / 1053.45
& 85.09 / 1274.22 \\
& KorBench
& 47.20 / 2731.22
& 44.80 / 2785.61
& 45.44 / 3792.71 \\
& IFBench
& 21.20 / 3404.26
& 19.73 / 3457.09
& 20.87 / 4769.71 \\
& MMLU-Pro
& 53.41 / 3235.44
& 53.62 / 3308.69
& 53.69 / 4552.05 \\
\midrule
\multicolumn{2}{l}{Average} 
& 55.58 / 2546.96
& 55.13 / 2590.62
& 55.01 / 3444.09 \\
\bottomrule
\end{tabular}}
\caption{Accuracy and throughput comparison for UpLoop-Ling-3.0 Tiny ($R=3$). We compare standard self-speculative decoding against two additional training-free cross-loop reuse strategies. Both sharing strategies preserve accuracy well overall, while reaching \textbf{2.30$\times$} speedup over non-speculative decoding.}
\label{tab:ling_moe_share_acc_tps}
\end{table*}

\paragraph{Validation of Routing Sharing and Recurrent-State Sharing.}
\Cref{tab:ling_moe_share_acc_tps} shows that both strategies maintain accuracy close to the self-speculative baseline while providing additional throughput gains.

For \textbf{Routing-Share$(3\leftarrow2)$}, the average accuracy is \textbf{55.13}, compared with \textbf{55.58} for the original self-speculative UpLoop, a small difference of only \textbf{-0.45}. 
Meanwhile, routing sharing raises average TPS from \textbf{1495.19} tokens/s in the non-speculative baseline to \textbf{2590.62} tokens/s, corresponding to a \textbf{1.73$\times$} speedup over non-speculative decoding. 
Relative to self-speculative decoding without sharing, it provides a modest gain.

For \textbf{Recurrent-Share$(3\leftarrow2)$}, the average accuracy is \textbf{55.01}, also close to the self-spec-only reference. 
At the same time, this strategy substantially improves system efficiency, raising average throughput to \textbf{3444.09} tokens/s~(  \textbf{2.30$\times$} over non-speculative decoding). 
For fairness, under recurrent-state sharing we increase the \texttt{mamba\_cache} from 1024 to 1536~(1.5x), since reusing the $r=2$ KDA recurrent stream at $r=3$ removes roughly one-third of the loop-specific recurrent-state storage for KDA layers. 
This gain indicates that, in Ling-3.0 Tiny, cross-loop recurrent-state reuse in UpLoop removes a large source of verification overhead.

\section{Conclusion}
\label{sec:conclusion}

We presented \textbf{UpLoop}, a post-training recipe that upcycles pretrained non-looped LLMs into recurrent-depth models via layer-wise looping and gated residual modulation. 
Starting from Qwen3-4B Base with SFT, UpLoop-Qwen3-4B improves the 20-benchmark average by +6.17\% at $R=3$, overcoming the commonly observed $R=2$ saturation bottleneck. 
While $R=3$ offers the optimal accuracy--cost trade-off, both native and extrapolated $R=4$ yield further gains. 
Moreover, UpLoop generalizes effectively beyond dense models to the MoE \textbf{Ling-3.0 Tiny}, boosting its 17-benchmark average by +4.54\%. 
Our analyses confirm that hidden-wise gating places repeated execution in a well-conditioned regime, turning later loops into controlled, progressive refinements. 
Finally, UpLoop supports practical inference acceleration beyond accuracy gains alone. 
Across both dense and MoE backbones, and across both GQA- and KDA-style architectures, we successfully adapt speculative decoding for looped inference, achieving up to \textbf{2.05$\times$} speedup on UpLoop-Qwen3-4B and \textbf{2.30$\times$} speedup on UpLoop-Ling-3.0 Tiny. 
Overall, UpLoop establishes gated layer-wise recurrence as both an effective and deployment-friendly post-training scaling paradigm for pretrained non-looped LLMs.

\clearpage

\section*{Acknowledgement}
This work was supported by the Ant Group Research Intern Program.

\bibliographystyle{antgroup}
\bibliography{reference}
\clearpage

\beginappendix

\section{Theoretical Perspective on Gated Recurrent Refinement}
\label{app:theory_gate}

This appendix formalizes the main theoretical claims in \Cref{sec:method_gate}. 
We study one looped physical layer in the abstract residual form
\begin{equation}
h_r = h_{r-1} + D_r F(h_{r-1}),
\qquad r=1,\dots,R,
\label{eq:app_dyn}
\end{equation}
where $h_0\in\mathbb{R}^d$ is the hidden state entering the layer, $F:\mathbb{R}^d\to\mathbb{R}^d$ is the effective residual transformation of that layer, and
\begin{equation}
D_1 = I_d,
\qquad
D_r = \operatorname{Diag}(g_r)\in[0,1]^{d\times d},
\quad r=2,\dots,R,
\label{eq:app_gate_diag}
\end{equation}
with $g_r\in[0,1]^d$ the hidden-wise gate vector for loop $r$. The ungated case corresponds to $D_r\equiv I_d$ for all $r$.

We make precise three points used in the main text:
(i) gating reduces worst-case perturbation amplification across loops,
(ii) in a local small-step regime, multi-loop execution admits a depth-like first-order expansion, and
(iii) the value of gating is not to make such an expansion possible in principle, but to make its assumptions more plausible in practice.

\subsection{Local Assumptions}
\label{app:gate_assumptions}

We work in a neighborhood $\mathcal U\subset\mathbb{R}^d$ containing the recurrent trajectory of interest and assume:

\begin{enumerate}[leftmargin=1.5em]
    \item \textbf{Local Lipschitzness.} There exists $L_F>0$ such that
    \begin{equation}
    \|F(x)-F(y)\|\le L_F\|x-y\|,
    \qquad \forall x,y\in\mathcal U.
    \label{eq:app_lipschitz_F}
    \end{equation}

    \item \textbf{Local smoothness.} $F$ is differentiable on $\mathcal U$ with bounded first derivatives. Equivalently, there exists $C_F>0$ such that for all $x\in\mathcal U$ sufficiently close to $h_0$,
    \begin{equation}
    \|F(x)-F(h_0)\| \le C_F\|x-h_0\|.
    \label{eq:app_local_smooth}
    \end{equation}

    \item \textbf{Bounded gate operators.} For each loop $r$, there exists $\alpha_r\in[0,1]$ such that
    \begin{equation}
    \|D_r\|_{\mathrm{op}} \le \alpha_r.
    \label{eq:app_D_bounds}
    \end{equation}
\end{enumerate}

These are local assumptions only. They are intended to describe the recurrent refinement trajectory actually visited during looping, not the global behavior of the full network.

\subsection{One-Step Stability}
\label{app:gate_stability_sec}

We first quantify how gating improves local stability by shrinking the residual contribution at each revisit.

\paragraph{Proposition 1 (one-step stability bound).}
Under \eqref{eq:app_lipschitz_F} and \eqref{eq:app_D_bounds}, the recurrent map
\begin{equation}
\Phi_r(h) := h + D_rF(h)
\end{equation}
satisfies
\begin{equation}
\|\Phi_r(x)-\Phi_r(y)\|
\le
\bigl(1+\alpha_rL_F\bigr)\|x-y\|,
\qquad \forall x,y\in\mathcal U.
\label{eq:app_phi_lipschitz}
\end{equation}
Consequently, for any two trajectories $\{h_r\}$ and $\{\tilde h_r\}$ that remain in $\mathcal U$,
\begin{equation}
\|h_R-\tilde h_R\|
\le
\Biggl(\prod_{r=1}^{R}(1+\alpha_rL_F)\Biggr)\|h_0-\tilde h_0\|.
\label{eq:app_trajectory_bound}
\end{equation}

\paragraph{Proof.}
For any $x,y\in\mathcal U$,
\begin{align}
\Phi_r(x)-\Phi_r(y)
&=
(x-y)+D_r(F(x)-F(y)).
\end{align}
Taking norms and applying the triangle inequality gives
\begin{equation}
\|\Phi_r(x)-\Phi_r(y)\|
\le
\|x-y\|+\|D_r\|_{\mathrm{op}}\|F(x)-F(y)\|.
\end{equation}
Using \eqref{eq:app_lipschitz_F} and \eqref{eq:app_D_bounds},
\begin{equation}
\|\Phi_r(x)-\Phi_r(y)\|
\le
\bigl(1+\alpha_rL_F\bigr)\|x-y\|,
\end{equation}
which proves \eqref{eq:app_phi_lipschitz}. Iterating over $r=1,\dots,R$ yields \eqref{eq:app_trajectory_bound}.
\hfill $\square$

\paragraph{Ungated comparison.}
If $D_r\equiv I_d$, then $\alpha_r=1$ and the one-step factor becomes
\begin{equation}
\|\Phi_r(x)-\Phi_r(y)\|\le (1+L_F)\|x-y\|.
\label{eq:app_ungated_lipschitz}
\end{equation}
Thus, whenever $\alpha_r<1$, gating strictly reduces the worst-case amplification contributed by the residual branch.

\paragraph{Interpretation.}
This is the local stability statement behind \eqref{eq:gated-one-step-bound} in the main text. Gating does not remove the residual transformation; it attenuates its worst-case contribution to perturbation growth, making repeated reuse better conditioned.

\subsection{First-Order Expansion in the Small-Step Regime}
\label{app:gate_firstorder_sec}

We next formalize the depth-like interpretation of looping. The key point is that if each gated write is small relative to the local hidden-state scale, then the trajectory remains near $h_0$, and repeated execution admits a controlled first-order expansion.

\paragraph{Proposition 2 (first-order depth expansion).}
Assume that the recurrent trajectory remains in $\mathcal U$ and that each gated residual step is uniformly small:
\begin{equation}
\|D_rF(h_{r-1})\|\le \varepsilon,
\qquad r=1,\dots,R.
\label{eq:app_small_step}
\end{equation}
Then
\begin{equation}
\|h_r-h_0\| \le r\varepsilon,
\qquad r=1,\dots,R.
\label{eq:app_hr_bound}
\end{equation}
Moreover, under \eqref{eq:app_local_smooth},
\begin{equation}
h_R
=
h_0
+
\sum_{r=1}^{R} D_rF(h_0)
+
\mathcal O(R^2\varepsilon^2).
\label{eq:app_first_order}
\end{equation}

\paragraph{Proof.}
From \eqref{eq:app_dyn},
\begin{equation}
h_r-h_{r-1}=D_rF(h_{r-1}).
\end{equation}
Hence, by repeated use of the triangle inequality and \eqref{eq:app_small_step},
\begin{equation}
\|h_r-h_0\|
\le
\sum_{t=1}^{r}\|h_t-h_{t-1}\|
=
\sum_{t=1}^{r}\|D_tF(h_{t-1})\|
\le
r\varepsilon,
\end{equation}
which proves \eqref{eq:app_hr_bound}.

Now write
\begin{equation}
F(h_{r-1}) = F(h_0) + \Delta_r,
\qquad
\Delta_r := F(h_{r-1})-F(h_0).
\label{eq:app_delta_def}
\end{equation}
By \eqref{eq:app_local_smooth} and \eqref{eq:app_hr_bound},
\begin{equation}
\|\Delta_r\|
\le
C_F\|h_{r-1}-h_0\|
\le
C_F(r-1)\varepsilon
=
\mathcal O(r\varepsilon).
\label{eq:app_delta_bound}
\end{equation}
Unrolling \eqref{eq:app_dyn},
\begin{equation}
h_R
=
h_0
+
\sum_{r=1}^{R}D_rF(h_{r-1}).
\label{eq:app_telescoping}
\end{equation}
Substituting \eqref{eq:app_delta_def} into \eqref{eq:app_telescoping},
\begin{align}
h_R
&=
h_0
+
\sum_{r=1}^{R}D_rF(h_0)
+
\sum_{r=1}^{R}D_r\Delta_r.
\end{align}
Using \eqref{eq:app_D_bounds} and \eqref{eq:app_delta_bound},
\begin{equation}
\|D_r\Delta_r\|
\le
\|D_r\|_{\mathrm{op}}\|\Delta_r\|
\le
\alpha_r\,\mathcal O(r\varepsilon)\,\varepsilon
=
\mathcal O(r\varepsilon^2),
\end{equation}
where the last step uses $\alpha_r\le 1$ and the fact that each residual step has size at most $\varepsilon$. Therefore,
\begin{equation}
\left\|
\sum_{r=1}^{R}D_r\Delta_r
\right\|
\le
\sum_{r=1}^{R}\mathcal O(r\varepsilon^2)
=
\mathcal O(R^2\varepsilon^2),
\end{equation}
which proves \eqref{eq:app_first_order}.
\hfill $\square$

\paragraph{Ungated comparison.}
If $D_r\equiv I_d$, the same argument yields
\begin{equation}
h_R
=
h_0
+
R\,F(h_0)
+
\mathcal O(R^2\varepsilon^2).
\label{eq:app_first_order_ungated}
\end{equation}
So the point is not that only gated recurrence admits such an expansion. Rather, gating replaces repeated full-strength residual accumulation with loop-dependent small-step accumulation, which makes the underlying local regime much more plausible in practice.

\paragraph{Interpretation.}
Equation~\eqref{eq:app_first_order} is the formal version of the main-text claim that a looped layer behaves like a depth-expanded residual stack to leading order. This is not equivalence to an arbitrary untied depth-$R$ network. The claim is narrower: in a local small-step regime, repeated execution decomposes into a shared residual accumulation term plus a controlled higher-order correction.

\paragraph{Connection to measurements.}
In the main experiments, later-round hidden updates are empirically small relative to the ambient hidden-state scale, and cumulative displacement remains modest across loops. These measurements support the local small-step regime assumed above and help explain why gated looping remains useful beyond $R=2$.

\subsection{Jacobian View}
\label{app:gate_jacobian_sec}

A complementary local view comes from the Jacobian of the recurrent map.

\paragraph{Proposition 3 (Jacobian control).}
Assume $F$ is differentiable on $\mathcal U$. Then
\begin{equation}
J_{\Phi_r}(h)=I_d + D_rJ_F(h).
\label{eq:app_jacobian_exact}
\end{equation}
Consequently,
\begin{equation}
\|J_{\Phi_r}(h)-I_d\|_{\mathrm{op}}
\le
\|D_r\|_{\mathrm{op}}\|J_F(h)\|_{\mathrm{op}}.
\label{eq:app_jacobian_bound}
\end{equation}
In particular, if $\|J_F(h)\|_{\mathrm{op}}\le M_F$ on $\mathcal U$, then
\begin{equation}
\|J_{\Phi_r}(h)-I_d\|_{\mathrm{op}}
\le
\alpha_r M_F.
\label{eq:app_jacobian_bound_simple}
\end{equation}

\paragraph{Proof.}
Differentiate $\Phi_r(h)=h+D_rF(h)$. Since $D_r$ is constant with respect to $h$,
\begin{equation}
J_{\Phi_r}(h)=I_d+D_rJ_F(h),
\end{equation}
which proves \eqref{eq:app_jacobian_exact}. Taking operator norms and using submultiplicativity gives \eqref{eq:app_jacobian_bound}; then \eqref{eq:app_D_bounds} gives \eqref{eq:app_jacobian_bound_simple}.
\hfill $\square$

\paragraph{Interpretation.}
The Jacobian bound shows that gating keeps each recurrent step closer to identity plus a controlled correction. This is the differential analogue of the one-step Lipschitz bound: smaller gated writes imply less abrupt distortion of local hidden-space geometry.

\subsection{Continuous-Depth Interpretation}
\label{app:gate_ode}

The recurrence in \eqref{eq:app_dyn} can also be viewed as a discretized residual flow,
\begin{equation}
\frac{dh(t)}{dt}=A(t)F(h(t)),
\label{eq:app_ode}
\end{equation}
where $A(t)$ is diagonal and positive semidefinite. If one discretizes time with step sizes $\Delta t_r$ and identifies
\begin{equation}
D_r \approx \Delta t_r A(t_r),
\label{eq:app_ode_identify}
\end{equation}
then \eqref{eq:app_dyn} becomes
\begin{equation}
h_r \approx h_{r-1} + \Delta t_r A(t_r)F(h_{r-1}),
\label{eq:app_euler}
\end{equation}
which is an explicit Euler step for \eqref{eq:app_ode}. In this view, the gate acts as a learned coordinate-wise step-size controller. This interpretation is most meaningful precisely in the same small-step regime used in \Cref{app:gate_firstorder_sec}.

\subsection{Discussion}
\label{app:gate_discussion}

The results above support a narrow but practically relevant conclusion for post-training upcycling. 
\Cref{app:gate_stability_sec} shows that gating reduces worst-case perturbation amplification across loops. 
\Cref{app:gate_firstorder_sec} shows that, in a local small-step regime, repeated execution admits a depth-like first-order expansion with shared weights and loop-dependent coefficients. 
\Cref{app:gate_jacobian_sec} shows that gating keeps each recurrent map closer to identity in local geometry.

Taken together, these results clarify the role of hidden-wise gating. It does not add new parametric knowledge, nor does it make a recurrent layer equivalent to an arbitrary untied deeper network. Its role is more specific: it places repeated layer reuse in a better-conditioned local regime, where additional loops are more likely to act as controlled residual refinement rather than repeated full-strength overwrite. This is precisely the regime needed for stable and useful post-training depth upcycling.

\end{document}